%% file: main_dmlr2023.tex
\documentclass{article}

\usepackage{microtype}
\usepackage{graphicx}
\usepackage{booktabs} 

\usepackage{hyperref}
\usepackage{times}

\usepackage[accepted]{icml2023}

\usepackage{amsmath}
\usepackage{amssymb}
\usepackage{mathtools}
\usepackage{amsthm}

\usepackage[capitalize,noabbrev]{cleveref}

\theoremstyle{plain}

\theoremstyle{definition}

\theoremstyle{remark}

\usepackage[textsize=tiny]{todonotes}

\usepackage [latin1]{inputenc}
\usepackage{url}

\usepackage{xcolor}
\usepackage{comment}
\usepackage{enumitem}
\usepackage{caption}
\usepackage{subcaption}
\usepackage{multirow}

\newcommand{\system}{WBA}
\newcommand{\systemlong}{weighted balanced accuracy}
\newcommand{\systemcap}{Weighted Balanced Accuracy}
\newcommand{\systemrare}{$\mathit{WBA_{rarity}}$}
\newcommand{\systemuser}{$\mathit{WBA_{user}}$}

\icmltitlerunning{A Skew-Sensitive Evaluation Framework for Imbalanced Data Classification}

\begin{document}

\twocolumn[
\icmltitle{A Skew-Sensitive Evaluation Framework for Imbalanced Data Classification}


\icmlsetsymbol{equal}{*}

\begin{icmlauthorlist}
\icmlauthor{Min Du}{equal,comp1}
\icmlauthor{Nesime Tatbul}{equal,comp2,u1}
\icmlauthor{Brian Rivers}{comp2}
\icmlauthor{Akhilesh Kumar Gupta}{comp5}
\icmlauthor{Lucas Hu}{comp1}
\icmlauthor{Wei Wang}{comp1}
\icmlauthor{Ryan Marcus}{u2}
\icmlauthor{Shengtian Zhou}{comp3}
\icmlauthor{Insup Lee}{u2}
\icmlauthor{Justin Gottschlich}{comp4,u3}
\end{icmlauthorlist}

\icmlaffiliation{u1}{MIT}
\icmlaffiliation{u2}{University of Pennsylvania}
\icmlaffiliation{u3}{Stanford University}
\icmlaffiliation{comp1}{Palo Alto Networks}
\icmlaffiliation{comp2}{Intel}
\icmlaffiliation{comp3}{Snap Inc.}
\icmlaffiliation{comp4}{Merly}
\icmlaffiliation{comp5}{Apple Inc.}

\icmlcorrespondingauthor{Min Du}{min.du.email@gmail.com}
\icmlcorrespondingauthor{Nesime Tatbul}{tatbul@csail.mit.edu}

\icmlkeywords{Imbalanced data classification, Evaluation metrics, Log parsing, Sentiment analysis, URL classification}

\vskip 0.3in
]



\printAffiliationsAndNotice{\icmlEqualContribution} 

\input{abstract}

\input{introduction}


\input{relatedwork}

\input{system}

\input{experiments}

\input{conclusion}


\bibliography{main_dmlr2023}
\bibliographystyle{icml2023}


\newpage
\appendix
\onecolumn
\input{appendix}


\end{document}

%% file: abstract.tex
\begin{abstract}


Class distribution skews in imbalanced datasets may lead to models with prediction bias towards majority classes, making fair assessment of classifiers a challenging task. Metrics such as Balanced Accuracy are commonly used to evaluate a classifier's prediction performance under such scenarios. However, these metrics fall short when classes vary in importance.
In this paper, we propose a simple and general-purpose evaluation framework for imbalanced data classification that is sensitive to arbitrary skews in class cardinalities and importances. Experiments with several state-of-the-art classifiers tested on real-world datasets from three different domains show
the effectiveness of our framework -- not only in evaluating and ranking classifiers, but also training them.

\end{abstract}

%% file: introduction.tex
\section{Introduction} \label{sec:intro}

\begin{figure*}[t]
\begin{center}
\subfloat[Skew in distributions]{
\includegraphics[width=0.37\linewidth]{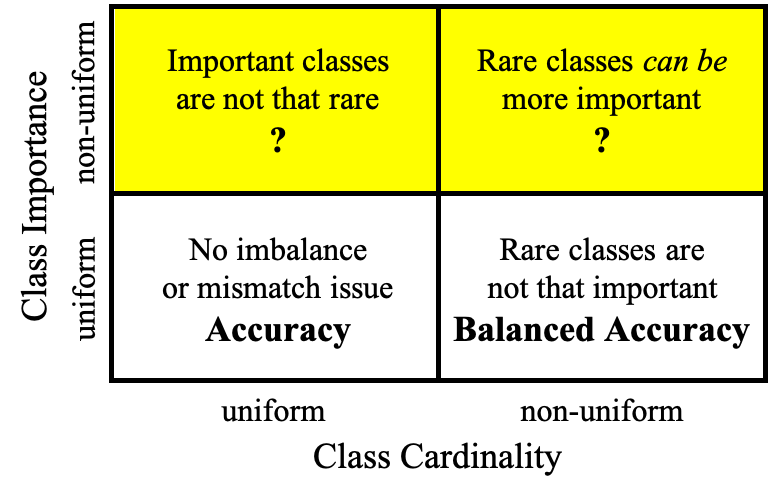}
\label{fig:quad-chart}
}
\subfloat[Example]{
\includegraphics[width=0.53\linewidth]{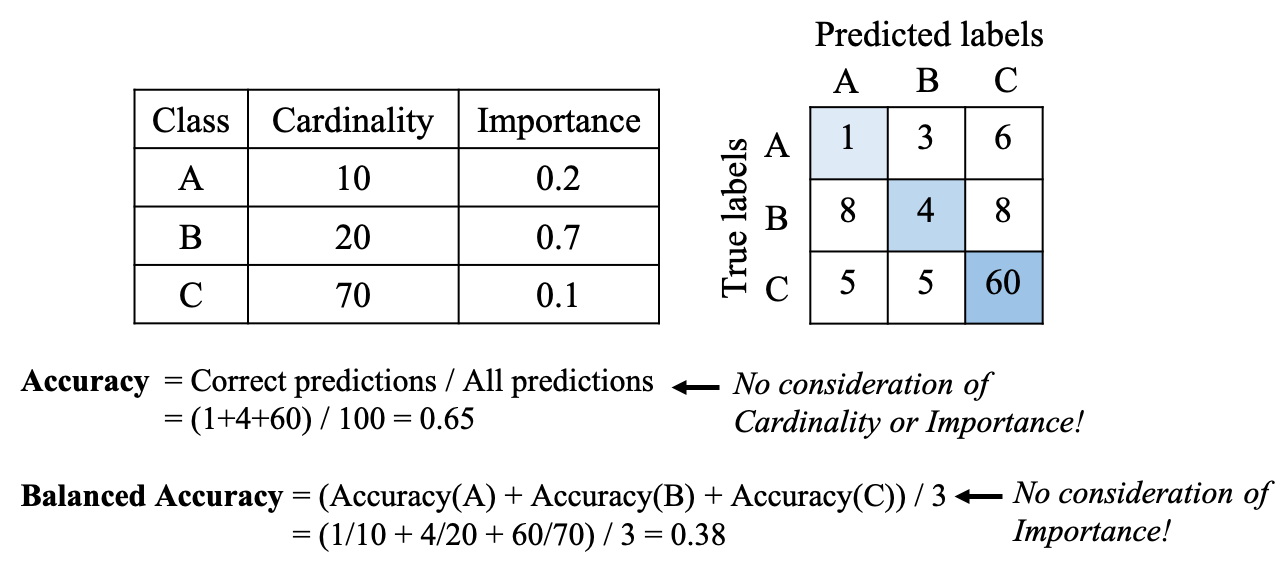}
\label{fig:example}
}
\caption{Skew in distributions of Class Cardinality or Class Importance, and the potential mismatch between them render existing accuracy metrics unusable in general multi-class prediction problems.}
\label{fig:skew}
\end{center}
\vspace{-5mm}
\end{figure*}

For a broad range of machine learning (ML) tasks, predictive modeling in the presence of {\em imbalanced datasets} -- those with severe distribution skews -- has been a long-standing problem~\citep{he:tkde09, sun:ijpr09, he13:book, branco:2016:survey, hilario18:book, dl-survey19}. Imbalanced training datasets lead to models with prediction bias towards majority classes, which in turn results in misclassification of the underrepresented ones. Yet, those minority classes often are the ones that correspond to the most important events of interest (e.g., errors in system logs~\citep{zhu:2019:icse}, infected patients in medical diagnosis~\citep{cohen:2006:cwa}, fraud in financial transactions~\citep{makki:access19}). While there is often an inverse correlation between the class cardinalities and their importance (i.e., rare classes are more important than others), the core problem here is the mismatch between the way these two distributions are skewed: the $i^{th}$ most common class is not necessarily the $i^{th}$ most important class (see Figure~\ref{fig:quad-chart} for an illustration). In fact, rarity is one of many potential criteria that can determine the importance of a class,
which is usually positively correlated with the costs or risks involved in its misprediction. Ignoring these criteria when dealing with imbalanced data classification may have detrimental consequences.

Consider automatic classification of messages in system event logs as an example \citep{zhu:2019:icse}. An \emph{event log} is a temporal sequence of messages that have transpired for a given software system (e.g., operating systems, cyber-physical systems) over a certain time period. Event logs are particularly useful after a system has been deployed, as they can provide the DevOps teams with insights about errors outside of the testing environment, thereby enabling them to debug and improve the system quality. There is typically an inverse correlation between the stability/maturity of a system and the frequency of the errors it produces in its event logs. Furthermore, the message types that appear least frequently in an event log are usually the ones with the greatest importance.
For example, an error message may indicate an event that leads to catastrophic system failure, which is expected to be rare but should be treated with great importance once it happens.

A plethora of approaches have been proposed for building balanced classifiers
\citep{sun:ijpr09, branco:2016:survey}. A fundamental issue that still remains an open challenge is the lack of a generally-accepted methodology for measuring classification performance. The traditional metrics, which are designed to evaluate average case performance (e.g., {\em Accuracy}) are not capable of correctly assessing the results in presence of arbitrary skew mismatches between class cardinalities and importances. On the other hand, metrics specifically proposed for imbalanced learning are either domain-specific, do not easily generalize beyond two classes, or can not support varying class importance (e.g., {\em Balanced Accuracy}) \citep{japkowicz13}.

Let us illustrate the problem with the simple example in Figure~\ref{fig:example}. The test dataset consists of 100 data items from 3 classes (A, B, C). The greatest majority of the items belong to class C (70), but class B (20) has the greatest importance (0.7). In other words, Cardinality and Importance are both non-uniform and in favor of different classes (i.e., representing the top-right quadrant of Figure~\ref{fig:quad-chart}). The confusion matrix on the right shows the results from a classifier run against this test dataset. Unsurprisingly, the classifier performed the best for the majority class C (60/70 correct predictions). When evaluated using the traditional {\em Accuracy} metric, neither Class Cardinality nor Class Importance is taken into account. If {\em Balanced Accuracy} is used instead, we observe the degrading impact of the Class Cardinality skew ($0.38 < 0.65$), but Class Importance is still not accounted for. This example demonstrates the need for a new evaluation approach that is sensitive to both Cardinality and Importance skew, as well as any arbitrary correlations between them. This is especially critical for ensuring a fair comparative assessment across multiple classifiers or problem instances.

Our goal in this paper is to design an evaluation framework for imbalanced data classification, which can be reliably used to measure, compare, train, and tune classifier performance
in a way that is sensitive to non-uniform class importance.
We identify two key design principles: 

\begin{itemize}
[nosep,leftmargin=1em,labelwidth=*,align=left]

\item {\em Simplicity:} It should be {\em intuitive} and {\em easy} to use and interpret.

\item {\em Generality:} It should be general-purpose, i.e., (i) {\em extensible} to an arbitrary number of classes and (ii) {\em customizable} to any application domain.

\end{itemize}

To meet the first goal, we focus on scalar metrics such as \emph{Accuracy} (as opposed to graphical metrics such as ROC curves), as they are simpler, more commonly used, and scale well with increasing numbers of classes and models. To meet the second goal, we target the more general $n$-ary classification problems (as opposed to binary), as well as providing the capability to flexibly adjust class weights to capture non-uniform importance criteria that may vary across application domains. Note that we primarily focus on {\em Accuracy} as our base scalar metric in this paper, as it is seen as the de facto metric for classification problems \citep{Scikit}. However, our framework is general enough to be extended to other scalar metrics, such as {\em Precision} and {\em Recall}. Similarly, while we deeply examine three use cases (log parsing, sentiment analysis, URL classification), our framework in principle is generally applicable to any domain with imbalanced class and importance distributions.

We first
provide a brief overview of related work in Section~\ref{sec:relatedwork}.
Section~\ref{sec:system} presents our new, skew-sensitive evaluation framework. In Section~\ref{sec:experiments}, we show the practical utility of our framework by applying it 
over:
(i) three log parsing systems (Drain~\citep{he:2017:drain}, MoLFI~\citep{messaoudi:2018:molfi}, Spell~\citep{du:2016:spell, du:2018:spell}) using four real-world benchmarks~\citep{zhu:2019:icse};
(ii) a variety of deep learning models developed for sentiment analysis on a customer reviews
dataset from Amazon \citep{ni:2019:ijcnlp}; and
(iii) an industrial use case for URL classification with real classifiers and datasets from four cyber-security companies.
Finally, we conclude in Section \ref{sec:conclusion}.

%% file: relatedwork.tex
\section{Related Work} \label{sec:relatedwork}

{\bf Imbalanced Data Classification.}
Imbalanced data is prevalent in almost every domain \citep{cohen:2006:cwa, batuwita:jbcb12, makki:access19}. The growing adoption of ML models in diverse application domains has led to a surge in imbalanced data classification research \citep{he:tkde09, sun:ijpr09, he13:book, branco:2016:survey, hilario18:book, dl-survey19}. While the techniques widely vary, they fall under four basic categories: pre-processing training data to establish balance via sampling techniques \citep{estabrooks04, blaszczynski15}, building custom learning techniques for imbalanced training data \citep{joshi01, castro13}, post-processing predictions from an imbalanced model \citep{maloof03}, and their hybrids \citep{estabrooks01}. In this paper, we do not propose a new imbalanced learning technique, but a general-purpose performance evaluation framework that could be used in the training and/or testing of models for any technique. Section \ref{sec:experiments} demonstrates the practical utility of our framework for a variety of real ML use cases.\\
{\bf Evaluation Metrics.}
Traditional metrics for evaluating prediction performance such as Accuracy, Sensitivity/Specificity (and their combination G-mean), Precision/Recall (and their combination F-Score) were not designed with imbalanced data issues in mind \citep{japkowicz13}. In fact, most of these were originally intended for binary classification problems. To extend them to more than 2 classes, macro-averaging (i.e., arithmetic mean over individual class measurements) is used. Macro-averaging treats classes equally \citep{branco:2016:survey}. Balanced Accuracy is a popular averaging-based approach. There are also probabilistic evaluation approaches that extend Balanced Accuracy with Bayesian inference techniques for both binary and multi-class problems \citep{brodersen:2010:balanced, carrillo:2014:balanced}. Close to our work, \citet{cohen:2006:cwa} introduced the notion of class weights, yet in the specific context of Sensitivity/Specificity for binary classification in the medical domain. Similarly, \citet{batuwita:jbcb12} proposed extensions to G-mean for the bio-informatics domain. In addition to these scalar (a.k.a., threshold) metrics, graphical (a.k.a., ranking) evaluation methods such as Receiver Operating Characteristic (ROC) curves or Precision-Recall (PR) curves (and the Area Under the Curve (AUC) for such curves) as well as their extensions to imbalanced data / multi-class problems were also investigated \citep{wauc08, japkowicz13}. While these methods provide more detailed insights into the operational space of classifiers as a whole, they do not easily scale with use in problems with a large number of classes \citep{branco:2016:survey}. 

%% file: system.tex
\section{Skew-Sensitive Evaluation Framework} \label{sec:system}

In this section, we present our new evaluation framework
for multi-class learning problems in presence of arbitrary skews among class distributions and/or importances. Our framework builds on and extends commonly used scalar / threshold metrics such as $Accuracy$. These metrics were originally designed for binary classification problems, where there is typically more emphasis on one class (the positive class, e.g., anomalies). To adopt them to multi-class problems where there is no such single-class emphasis, each class' metric can be computed separately and then an overall aggregation (i.e., arithmetic mean) can be performed. For example, $Accuracy$ has been extended to $Balanced Accuracy$ by following this approach. In our framework, we follow a similar aggregation strategy, however, we do it in a more generalized way that allows custom class weights to capture class importance. Furthermore, these class weights can be based on any importance criteria such as rarity, cost, risk, expected benefits, and possibly a hybrid of multiple such criteria. Therefore, it is critical to provide a flexible formulation that allows users or domain experts to adjust the weights as needed by their problem instance. In what follows, we present our new skew-sensitive evaluation framework in a top-down fashion. Using the basic notation summarized in Table \ref{tab:notation}, we first formulate the general framework, and then we describe how this framework can be customized to different importance criteria scenarios by specializing the weights in a principled manner. For ease of exposition, we first focus on $Accuracy$ as the underlying performance metric, and then we discuss how our approach can be adopted to other similar metrics. Finally, we end this section with a brief discussion of how our framework can be used in model training.

\vspace{-0.1cm}
\subsection{\systemcap\ (\system)} \label{sec:wba}
\vspace{-0.1cm}

\begin{table}[t]
\centering
\small
\caption{Notation}
\scalebox{0.9}{
\begin{tabular}{|c|l|}
\hline
{\bf Notation} & {\bf Description } \\
\hline
\hline
$N$ & total number of data items \\
\hline
$C$ & total number of data item classes \\
\hline
$M$ & number of importance criteria \\
\hline
$n_i$ & true number of data items in class $i$ \\
\hline
$p_i$ & correctly predicted number of data items in class $i$ \\
\hline
$f_i$ & relative frequency of class $i$ \\
\hline
$w_i$ & relative weight of class $i$ \\
\hline
$u_i$ & relative user-defined importance of class $i$ \\
\hline
$r_i$ & relative rarity of class $i$ \\
\hline
$m_{i,j}$ & relative weight of class $i$ for importance criteria $j$ \\
\hline
${Accuracy}_i$ & Accuracy of class $i$ \\
\hline
\end{tabular}
}
\label{tab:notation}
\vspace{-3mm}
\end{table}

Suppose we are given a test dataset with $N$ data items in it, each of which belongs to one of $C$ distinct classes. Furthermore, each class $i$ contains $n_i$ of the data items in this dataset. Thus: $N = \sum_{i=1}^{C}{n_i}$.


The relative frequency of each class $i$ in the whole dataset: 

\vspace{-0.3cm}
\begin{footnotesize}
\begin{equation}
f_i = \frac{n_i}{N}
\label{eq:f}
\end{equation}
\end{footnotesize}
\vspace{-0.5cm}

Assume a classifier that makes a prediction about the class label of each data item in the test dataset, and correctly predicts $p_i$ out of $n_i$ labels for a given class $i$, where $p_i \leq n_i$. Then, the total number of correct predictions out of all the predictions gives us the overall $Accuracy$ of the classifier: 

\vspace{-0.3cm}
\begin{footnotesize}
\begin{equation}
Accuracy = \frac{\sum_{i=1}^{C}{p_i}}{N} 
\label{eq:acc}
\end{equation}
\end{footnotesize}
\vspace{-0.5cm}

The classifier's $Accuracy_i$ for a given class $i$ (a.k.a., per-class $Recall$ score) can be computed as: ${Accuracy}_i = \frac{p_i}{n_i} $.


$BalancedAccuracy$ is the macro-average of
${Accuracy}_i$
over all classes in the dataset:

\vspace{-0.3cm}
\begin{footnotesize}
\begin{equation}
Balanced Accuracy = \frac{1}{C} \times \sum_{i=1}^{C}{{Accuracy}_i}
\label{eq:bal-acc}
\end{equation}
\end{footnotesize}
\vspace{-0.5cm}

The above formulation represents the state of the art in how prediction accuracy is evaluated for multi-class classifiers in presence of imbalanced datasets (i.e., those where $f_i$ are not even). While for balanced datasets (i.e., $\forall i, n_i = N/C \textrm{~and~} f_i = 1/C$)
$BalancedAccuracy = Accuracy$, for imbalanced datasets, $BalancedAccuracy$ ensures that the prediction accuracy is not inflated due to high-frequency classes' results dominating over the others'. $BalancedAccuracy$ works well as long as each class is of the same importance, since it is the simple arithmetic mean across per-class accuracy measurements (i.e., each class' accuracy contributes evenly to the overall accuracy). As we discussed in earlier sections, in many real-world classification problems, this assumption does not hold. Rather, classifiers must be rewarded higher scores for their prediction performance on more important classes. In order to capture this requirement, we generalize $BalancedAccuracy$ into $WeightedBalancedAccuracy$ by extending it with per-class importance weights $w_i$ as follows:

\vspace{-0.7cm}
\begin{footnotesize}
\begin{equation}
Weighted Balanced Accuracy = \sum_{i=1}^{C}{w_i \times {Accuracy}_i}
\label{eq:wbal-acc}
\end{equation}
\end{footnotesize}
\vspace{-0.6cm}

This simple yet powerful extension enables us to capture both skews and imbalances in class cardinalities as well as importances (i.e., the complete design space in Figure \ref{fig:quad-chart}). This general formulation can support any importance criteria for weights as long as $0 \leq w_i \leq 1 \textrm{~and~} \sum_{i=1}^{C}{w_i} = 1$. In the following subsections, we present
general use of WBA with custom weights, for other scalar metrics, as well as in improving model training.

\vspace{-0.1cm}
\subsection{Weight Customization} \label{sec:weight}
\vspace{-0.1cm}

In a multi-class setting, not only may the classes carry different importance weights, but also the criteria of importance may vary from one problem or domain to another. We now discuss several types of criteria that we think are commonly seen in applications. This is not meant to be an exhaustive list, but it provides examples and templates that can be easily tailored to different problems.

\noindent
{\bf Importance criteria = User-defined.} This is the most general and flexible form of importance criteria. The application designer or domain expert specifies the relative weight of each class based on some application-specific criteria. As an example, the problem might be about classifying images of different types of objects in highway traffic and the user gives higher importance to correct recognition of certain objects of interest (e.g., pedestrians, bikes, animals, etc).
We express user-defined relative weight of a class $i$ with $u_i$, which is simply used as $w_i$ in Equation \ref{eq:wbal-acc} (i.e., $w_i = u_i$).


\noindent
{\bf Importance criteria = Rarity.}
It is often the case that the rarer something is, the more noteworthy or valuable it is. In multi-class problems, this corresponds to the case when importance of a class $i$ is inversely correlated with its relative frequency of occurrence ($f_i$) in the dataset. For example, in system log monitoring, log messages for more rarely occurring errors or exceptions (e.g., denial of service attack) are typically of higher importance. Therefore, a classifier that performs well on detecting such messages must be rewarded accordingly. In our framework, we capture rarity using weights that are based on normalized inverse class frequencies formulated as follows:

\vspace{-0.3cm}
\begin{footnotesize}
\begin{equation}
w_i = r_i = \frac{1}{f_i \times \sum_{j=1}^{C}{\frac{1}{f_j}}}
\label{eq:w-rarity}
\end{equation}
\end{footnotesize}
\vspace{-0.5cm}

\noindent
{\bf Multiple importance criteria.}
In some problems, importance of a class depends on multiple different criteria (e.g., both rarity and a user-defined criteria). To express class weights in such scenarios, we leverage techniques from multi-criteria decision making and multi-objective optimization \citep{mcdm00, helff16}. One of the most basic methods is using normalized weighted sums based on composite weights \citep{helff16}. Composite weights can be computed either in additive or multiplicative form \citep{tofallis14}. The multiplicative approach tends to promote weight combinations that are uniformly higher across all criteria, and as such is found to be a more preferred approach in application scenarios similar to ours \citep{helff16, tofallis14}. While we present this approach here, in principle, other approaches from multi-criteria decision making theory could also be used within our framework. Given $M$ different criteria with $m_{i,j}$ denoting the relative weight of class $i$ for criteria $j$, we can compute the composite weight of a class $i$ as follows:

\vspace{-0.5cm}
\begin{footnotesize}
\begin{equation}
w_i = \frac{\prod_{j=1}^{M}{m_{i,j}}}{\sum_{k=1}^{C}{\prod_{j=1}^{M}{m_{k,j}}}}
\label{eq:w-composite}
\end{equation}
\end{footnotesize}
\vspace{-0.5cm}

For example, if we had two criteria, rarity $r$ and user-defined $u$ with weights $r_i$ and $u_i$ for each class $i$, respectively, then the composite weight for class $i$ would be $w_i = \frac{r_i \times u_i}{\sum_{j=1}^{C}{r_j \times u_j}}$.

\noindent
{\bf Partially-defined importance criteria.}
One commonly expected scenario (especially in those classification problems where the number of classes $C$ can be very large) is that not all of the class importance weights might be supplied by the user. For example, in a sentiment analysis use case, the user supplies the weights for all the negative classes, and leaves the others unspecified. Our framework can support such cases by automatically assigning weights to the unspecified classes. The default approach is to distribute the remaining portion of weights evenly across all unspecified classes: (1 - total weights specified) / (number of unspecified classes). If the user prefers an alternative approach (e.g., distribute the remainder based on rarity of the unspecified classes), this can also be easily supported by our framework.

\vspace{-0.1cm}
\subsection{Metric Customization} \label{sec:metric}
\vspace{-0.1cm}

The skew-sensitive evaluation framework presented above focused on the popular {\em Accuracy} metric as the underlying metric of prediction performance. However, our framework follows a general structure based on the idea of {\em weighted macro-averaging with customizable weights}, which can essentially be used with any performance metric that can be computed on the basis of a class. For example, the macro-averaging approaches that are already being used for {\em Precision}, {\em Recall}, and {\em F-Score} could easily be extended with our customizable weighing approach by replacing {\em Accuracy} in our formulas with one of these metrics.

\vspace{-0.1cm}
\subsection{Model Training Improvement using Class Weights} \label{sec:training}
\vspace{-0.1cm}

The customized weights presented in Section~\ref{sec:weight} not only helps to give a user-preferred ranking via the proposed WBA metric, but also helps to improve model training.
Recall that ML model training aims to minimize the loss between model predictions and ground truth labels. A common practice is to minimize the sum of all per-sample losses. Using $Loss_i$ to denote the total loss incurred by all samples within class $i$, the model loss to minimize in training would be:
$\mathcal{L} = \sum_{i=1}^{C}{{Loss}_i}$.

By applying class importance weights $w_i$ as suggested in Section~\ref{sec:weight}, the loss value for important classes takes a larger portion in the final loss value to minimize. This enables the back-propagation process to focus more on optimizing the model parameters for the higher weighted classes, and thus improves their accuracy. 
This way, the model loss becomes:
$\mathcal{L} = \sum_{i=1}^{C}{w_i \times {Loss}_i}$.


The advancement of popular deep learning frameworks have made this process rather straightforward to implement. For example, the TensorFlow API provides a parameter named {\tt class\_weight} 
to allow
user to pass in a JSON data structure that specifies a weight value for each class label (\cite{tf-class-weights}). 
Similarly, the PyTorch API for cross entropy loss also provides a parameter to pass in class weights (\cite{pytorch-class-weights}).
These provide perfect interfaces for the usage of the importance weights in our WBA framework, both to improve a model's learning ability of the important classes 
and raise
the overall WBA score of the classification outcomes.

%% file: experiments.tex
\vspace{-0.1cm}
\section{Experimental Study} \label{sec:experiments}
\vspace{-0.1cm}

We now present an experimental analysis of our new framework for three application domains. Our goal is to demonstrate the value of {\system} compared to existing metrics when evaluating ML models over real-world imbalanced data classification problems. As we will show, often times a traditional metric like $Accuracy$ or $BalancedAccuracy$ will make classifier $A$ seem preferable to classifier $B$, when in reality classifier $B$ is superior. In addition, we also provide an analysis of how \system\ can positively impact, not only the testing of models, but also their training. Details of our experiments (incl. code, data, and examples) can be found on GitHub\footnote{https://github.com/2023-07-03/weighted-balanced-accuracy} and in the appendix.

\vspace{-0.1cm}
\subsection{Use Case 1: Learned Log Parsing} \label{sec:log-exp}
\vspace{-0.1cm}

\begin{figure*}[!t]
    \begin{center}
        \subfloat[{ macOS (skew = 8.454)}]{
        \includegraphics[width=0.4\linewidth]{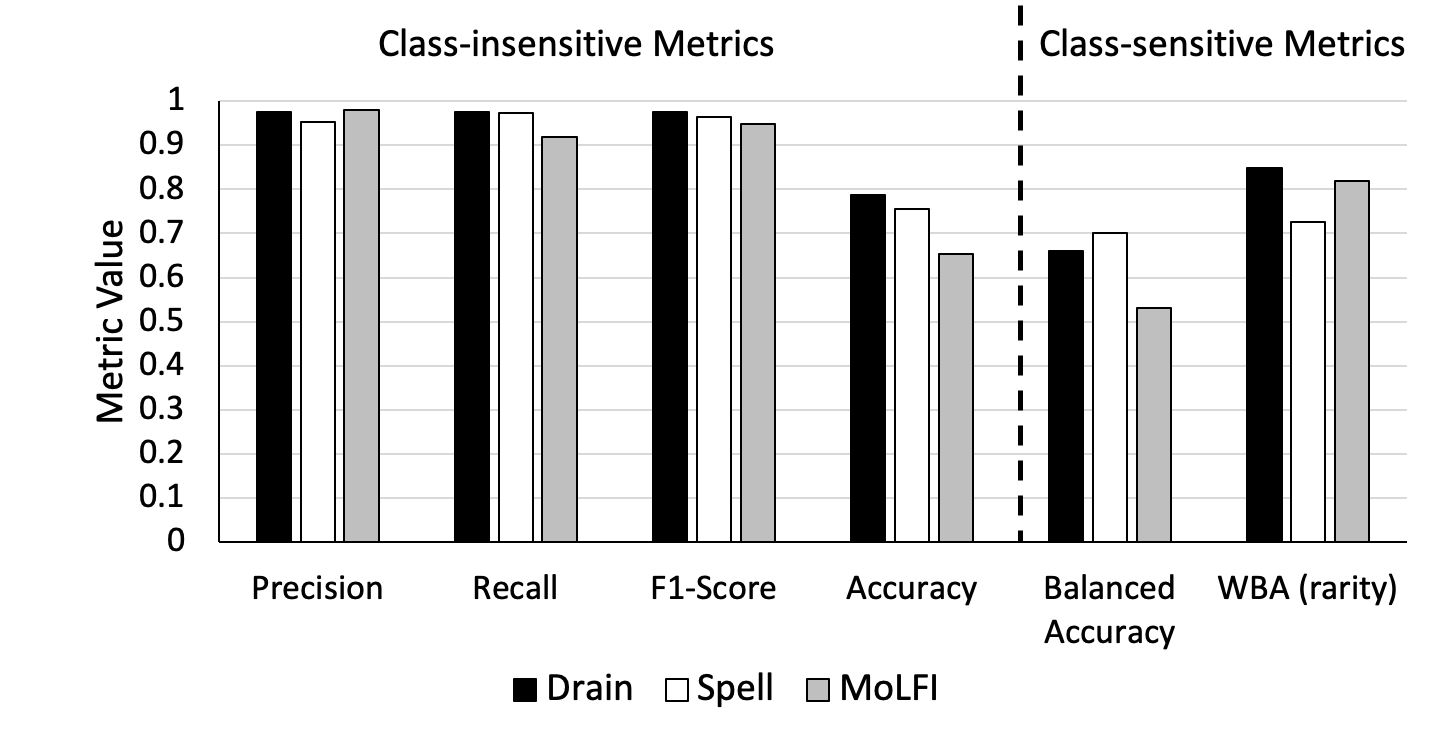}
        \label{fig:res-macos}
        }
        \subfloat[{ BGL (skew = 8.900)}]{
        \includegraphics[width=0.4\linewidth]{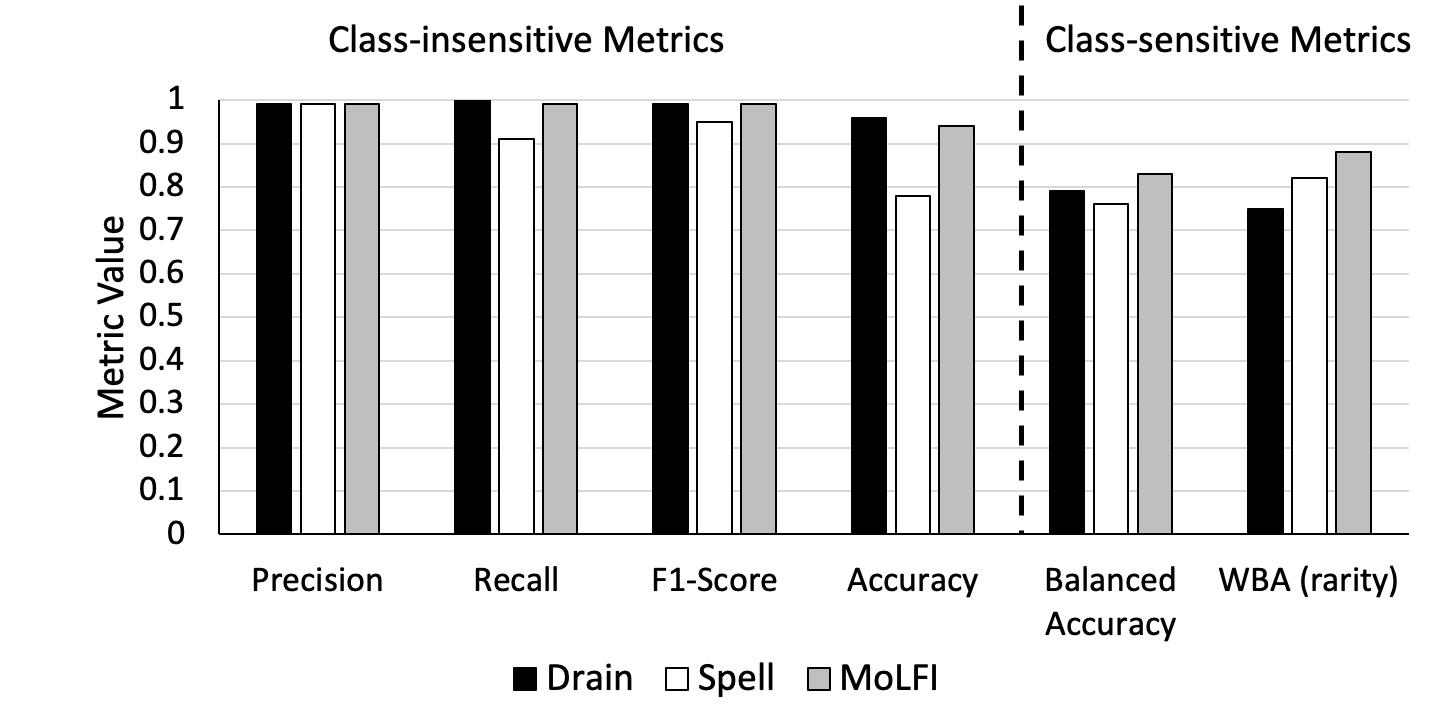}
        \label{fig:res-bgl}
        }
        \\
        \subfloat[{ Android (skew = 4.822)}]{
        \includegraphics[width=0.4\linewidth]{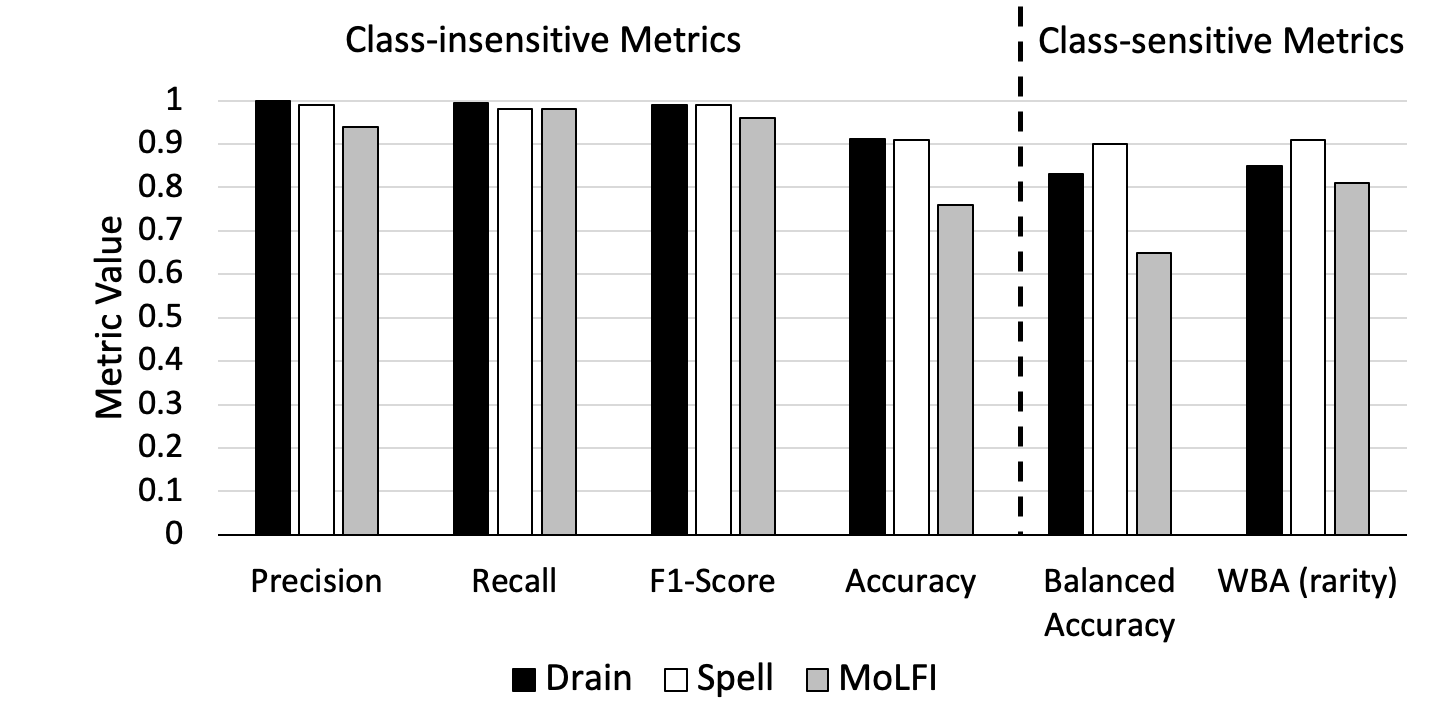}
        \label{fig:res-android}
        }
        \subfloat[{ HDFS (skew = 0.202)}]{
        \includegraphics[width=0.4\linewidth]{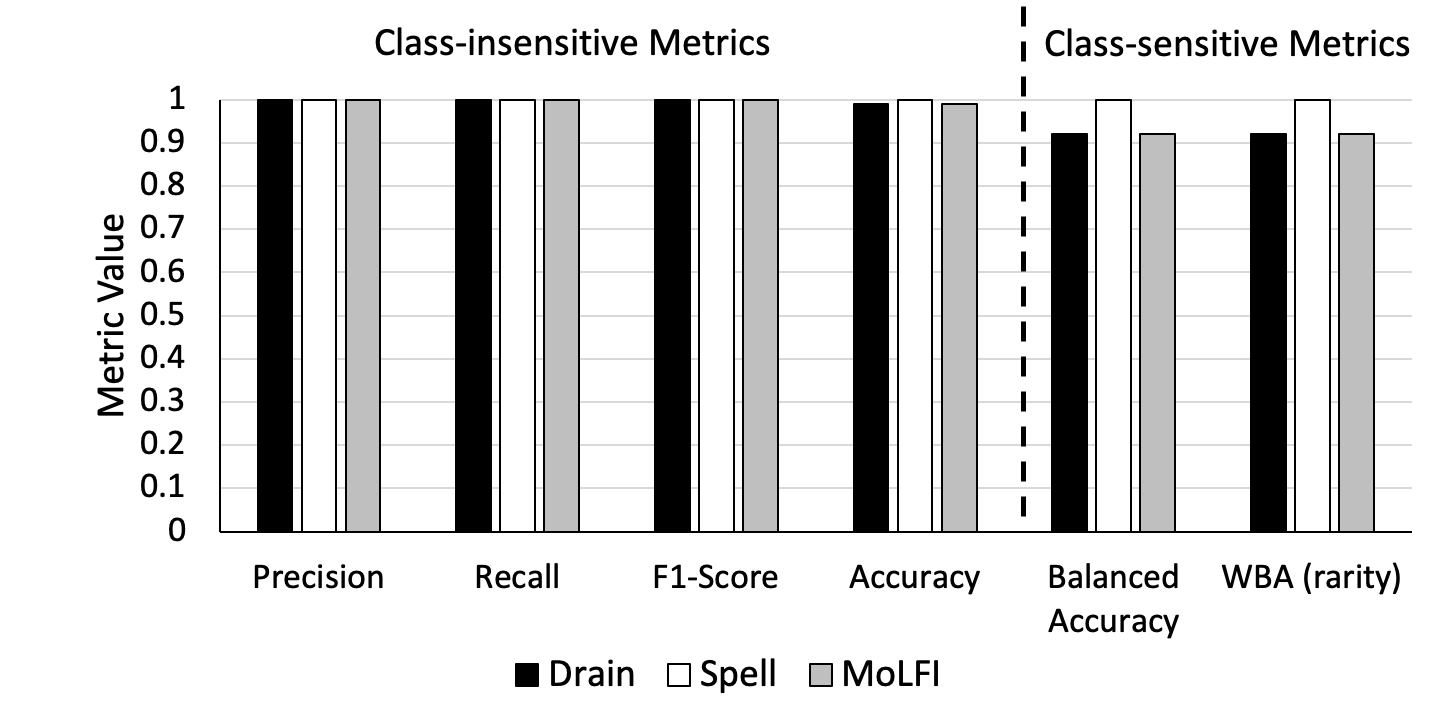}
        \label{fig:res-hdfs}
        }
        \caption{\system\ vs. class-insensitive \& class-sensitive metrics for log parsing: F1-Score \& Accuracy agree in all. BA \& WBA agree in (c) \& (d) only. WBA disagrees with class-insensitive in all.}
        \label{fig:exp-log}
        \vspace{-6mm}
    \end{center}
\end{figure*}

ML-based log parsers are tools that are designed to automatically learn the structure of event logs generated by hardware and software systems to properly categorize them into event classes (e.g., different error types). In our first study, we used \system\ to evaluate 3 state-of-the-art log systems: Drain, Spell, and MoLFI \citep{du:2016:spell, he:2017:drain, messaoudi:2018:molfi}. We start by providing an abbreviated description of our experimental setup. \\
\textbf{Log Parsing Systems.} \emph{Drain} is a rule-based, online log parsing system that encodes the parsing rules in a fixed-depth parse tree~\citep{he:2017:drain}. It performs a pre-processing step for each new log message using regular expressions created by domain experts. \emph{Spell}, 
on the other hand,
principally uses 
\emph{longest common subsequence} (LCS) to find new log message classes ~\citep{du:2016:spell}.
Finally, \emph{MoLFI} casts log parsing as a multi-objective optimization problem and provides a solution based on genetic programming~\citep{messaoudi:2018:molfi}. \\
\textbf{Datasets.} We test each aforementioned log message classification system with four real-world datasets taken from a public benchmark~\citep{zhu:2019:icse}. Each dataset has 2000 log instances randomly sampled from a larger dataset. 
The \emph{macOS} dataset contains raw log data generated by the macOS operating system (341 log classes, 237 {\em infrequent classes} (i.e., those that have fewer occurrences in the dataset than the average number of messages per class),
and an average class frequency of 5). The \emph{BlueGene/L (BGL)} dataset is a collection of logs from the BlueGene/L supercomputer system (120 log classes, 101 infrequent classes, and an average class frequency of 16). The \emph{Android}
dataset consists of logs from the Android mobile operating system~\citep{zhu:2019:icse} (166 log classes, 127 infrequent classes, and an average class frequency of 16). Finally, the \emph{HDFS}
dataset consists of log data collected from the Hadoop Distributed File System (14 log classes, 8 infrequent classes, and an average class frequency of 142). Overall, the first three datasets are highly skewed in class frequencies, whereas the HDFS dataset is relatively much less skewed (see Appendix \ref{sec:appendix-log}). \\
\textbf{Results.}
For Drain, Spell, and MoLFI, traditional metrics of $Precision$, $Recall$, {\em F1-Score}, and $Accuracy$ (named {\em Parsing Accuracy} in the original papers) were used for training and testing classification performance. None of these metrics are class-sensitive, while in log parsing, messages have in fact varying importance across the classes. The importance criteria is rarity: the more rare an error message is, the more important it is to correctly classify this message. To capture this,
we configure the \system\ to \systemrare, which automatically assigns weights to \system\ based on the dataset classes' \emph{inverse frequencies}, as described in Section~\ref{sec:weight}.
Then we evaluate the test results from the 3 parsers over 4 datasets using \systemrare\ and compare against traditional metrics in two categories: class-insensitive and class-sensitive, as shown in Figure \ref{fig:exp-log}. \\
\underline{\emph{{\systemrare} vs. Class-insensitive Metrics:}}
The class-insensitive metrics (specifically, F1-Score and Accuracy) agree on how to rank the classification performance of the 3 parsers across all datasets (for macOS and Android, Drain $>$ Spell $>$ MoLFI; for BGL, Drain $>$ MoLFI $>$ Spell; for HDFS, all perform similarly). Since \systemrare\ is sensitive to classes' data distribution and importance skews, it makes a completely different judgement. Furthermore, it ranks the techniques differently for each dataset (Drain $>$ MoLFI $>$ Spell in macOS; 
MoLFI  $>$ Spell $>$ Drain for BGL;
Spell $>$ Drain $>$ MoLFI for Android; and for HDFS, Spell $>$ Drain and MoLFI). 
The {\systemrare} ranking aligns with our observation on per-class accuracy of the methods. For example, on BGL dataset, although Spell has the most mis-classified samples and in most classes (and thus the lowest overall Accuracy and Balanced Accuracy), it has very few mis-classified samples for rare classes, making its {\systemrare} higher than Drain. On the other hand, MoLFI performs the best on rare classes and thus has the best {\systemrare}. A discussion on per-class performance can be found
in Appendix \ref{sec:appendix-log}.
This result validates that \systemrare\ provides a more sensitive tool for assessing classification performance. \\
\underline{\emph{{\systemrare} vs. Balanced Accuracy (BA):}}
As discussed earlier, BA is class-sensitive, but only to distribution imbalance. We can observe the difference between BA and WBA in Figure~\ref{fig:exp-log}. In macOS and BGL, where the skew is the highest and rarity is more pronounced, the two metrics completely disagree in how they rank the parsers. In contrast, for Android and HDFS, where the skew is lower, there is an overall agreement, although the separation in metric values slightly differ.
Of particular importance is the difference seen in Figure \ref{fig:res-macos}. We observe that the best performing model is Spell when scored by BA, and Drain when scored by {\systemrare}. The reason for this difference is due to 
the
differences in their ability to correctly classify infrequent classes, i.e., those that represent failures and errors that require the most immediate response.

\vspace{-0.1cm}
\subsection{Use Case 2: Sentiment Analysis} \label{subsec:amazon}
\vspace{-0.2cm}
\begin{table}[htb]
\centering
\caption{Amazon per-class breakdown: Frequencies $f_i$ are highly skewed (skew=2.140); $Accuracy_{i}$ in each model when both trained+tested with user-defined weights $w_i$ (same weights as in Figure \ref{fig:wba-amazon}).}
\label{tab:stats-amazon}
\vspace{-2mm}
{\scriptsize
\begin{tabular}{|c|c|c|c|c|c|c|}
\hline
{\bf Class} & {\bf Frequency} & {\bf Weights} & {\bf LSTM} & {\bf RNN} & {\bf GRU} & {\bf BiLSTM} \\
\hline
\hline
1 & 0.092 & 0.7 & 0.19 & 0.04 & 0.16 & 0.17 \\
\hline
2 & 0.052 & 0 & 0 & 0 & 0 & 0 \\
\hline
3 & 0.075 & 0 & 0 & 0 & 0 & 0 \\
\hline
4 & 0.142 & 0 & 0 & 0 & 0 & 0 \\
\hline
5 & 0.639 & 0.3 & 0.81 & 0.96 & 0.84 & 0.83 \\
\hline
\end{tabular}
}
\vspace{-2mm}
\end{table}

\begin{figure*}[t]
\begin{center}
\subfloat[WBA vs. Other Metrics (Train+Test)]{
\includegraphics[width=0.33\linewidth]{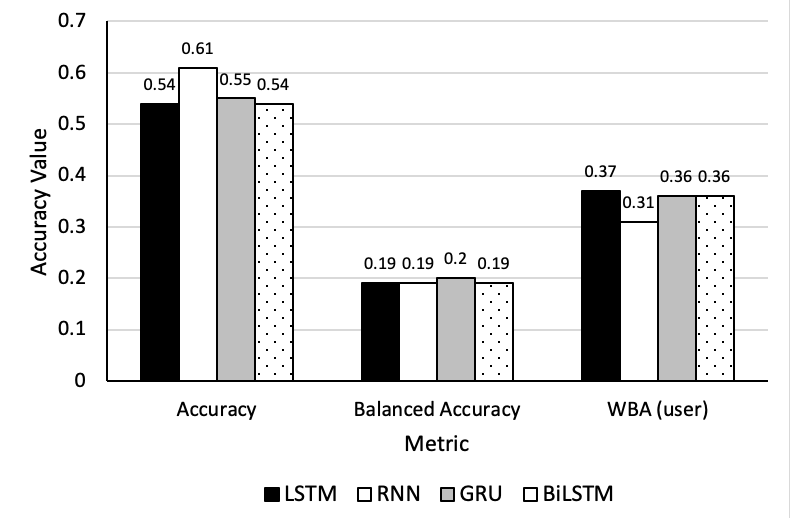}
\label{fig:wba-amazon}
}
\subfloat[WBA: Test vs. Train+Test]{
\includegraphics[width=0.41\linewidth]{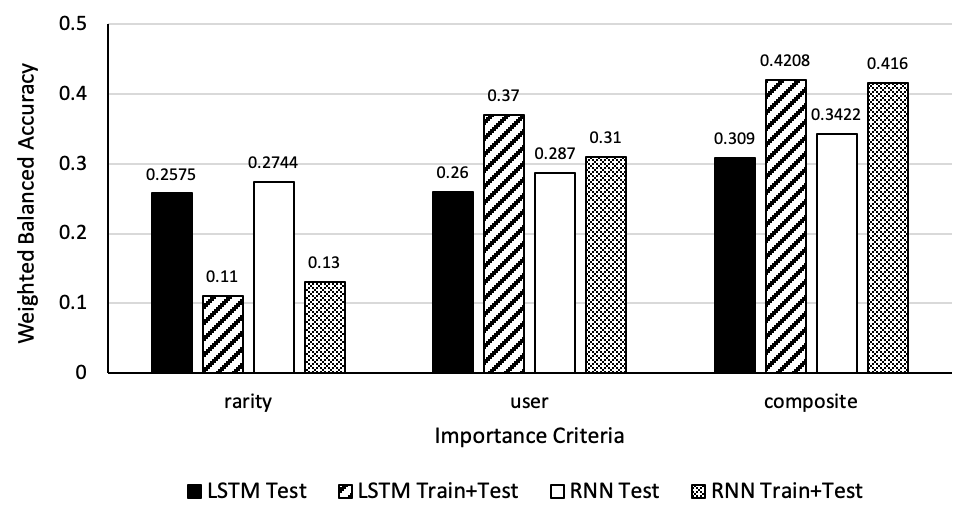}
\label{fig:train-amazon}
}
\vspace{-2mm}
\caption{Amazon results}
\label{fig:exp-amazon}
\vspace{-6mm}
\end{center}
\end{figure*}

In social media and other user-facing domains like e-commerce sites, it is often  useful to understand the view or feelings (``sentiments") associated with users' behavior or preferences. In the second part of our experimental study, we apply \system\ in the context of such a sentiment analysis use case, which involves analyzing text-based product reviews from Amazon's e-commerce websites. \\
\textbf{Dataset.}
The dataset consists of customer reviews and ratings of Amazon products
\citep{ni:2019:ijcnlp}. The task is to classify the reviews into 5 classes (with 1 being the lowest and 5 being the highest review rating a product can get), where ratings constitute the ground truth class labels. There is high class imbalance in this dataset (skew=2.140). As shown in the Frequency column of Table \ref{tab:stats-amazon}, Class 5 with the highest customer rating clearly dominates compared to the others. It is known that the distribution of customer review ratings is typically imbalanced and generally follow a J-shaped distribution~\citep{mudambi:2010:amazon, pavlou:2006:informs}. \\
\textbf{Sentiment Analysis Models.}
We compare 4 types of recurrent neural networks (RNN), all consisting of an embedding layer with pre-trained word embeddings from \citep{pennington:2014:glove} followed by a recurrent layer from PyTorch \citep{subramanian:2018:pytorch}: RNN, LSTM, GRU, BiLSTM. The hidden state output from the last time step of these are passed to a fully-connected layer with input of 256 neurons and output from 5 neurons. \\
\textbf{Results.}
For this use case, we first worked with a user-defined importance criteria borrowed from published studies suggesting that extreme review ratings (classes 1 and 5) carry more importance ~\citep{mudambi:2010:amazon, pavlou:2006:informs}. Thus, we set the weights as 
in Table \ref{tab:stats-amazon}
(shown as WBA(user) or user in Figure \ref{fig:exp-amazon}). \\
\underline{\emph{WBA vs. Other Accuracy Metrics:}}
First, we compare WBA(user) with Accuracy and BalancedAccuracy (BA) when used as a metric for both training and testing of the 4 DNN models (Figure \ref{fig:wba-amazon}). We make a few observations: (i) The class-insensitive Accuracy showcases the imbalance problem in classification, as it favors the RNN model which is heavily biased by the majority class (see $Accuracy_i$ for RNN in Table \ref{tab:stats-amazon} where class 5 scores 0.96). (ii) The frequency-sensitive BA metric finds all models perform similarly.
WBA(user), in contrast, identifies LSTM as the best model. Indeed, Table \ref{tab:stats-amazon} confirms that LSTM performs relatively the best in predicting the most important class, class 1 (0.19 accuracy). Overall, we find that WBA is capable of capturing importance skews, even when the frequency skew can be high and biased towards less important classes. \\
\underline{\emph{Impact of WBA in Model Training:}}
Next we explore the use of WBA not only in model evaluation, but also in training. We focus on two models (LSTM and RNN), and apply WBA only during testing vs. to both training (by extending loss functions of DNNs with weights as in Section \ref{sec:training}) and testing. Intuitively, if a model is trained being aware of the importance weights, then it should also perform well when tested against the same criteria. To test this hypothesis, we repeated the experiment for 3 alternative importance criteria: (i) rarity ($w_1 = 0.209$, $w_2 = 0.368$, $w_3 = 0.255$, $w_4 = 0.136$, $w_5 = 0.030$), (ii) user-defined (i.e., with weights in Table \ref{tab:stats-amazon}), and (iii) composite of the two ($w_1 = 0.62$, $w_2 = w_3 = w_4 = 0$, $w_5 = 0.38$). In Figure \ref{fig:train-amazon}, we observe:
(i) Except for rarity, WBA for both LSTM and RNN improves when integrated into model training. This verifies our intuition, and shows that WBA is a useful metric not only for evaluation, but also for training.
(ii) When we zoom into rarity, we see that although class 2 is the most important, per-class accuracy for class 5 is much higher for both LSTM and RNN in the {\em Test-only} case, because both models are still trained heavily biased towards the majority class (5).
(iii) Though rarity by itself is not useful in training, when combined with user importance, it visibly improves the WBA scores. This shows that our multi-criteria composition approach is capable of combining importance criteria as intended.

\input{url}

%% file: url.tex
\begin{table*}[tb]
\caption{Evaluating and ranking the URL classification services}
\label{tab:url-metrics}
\vspace{-2mm}
\centering
\scriptsize
\begin{tabular}{|l|r|r||r|r|r|r|r|c|}
\hline
\multirow{2}{*}{\bf Category} & \multirow{2}{*}{\bf \#URLs}  & \multirow{2}{*}{\bf Rarity $w_i$} & \multirow{2}{*}{\bf User $w_i$} & \multicolumn{4}{|c|}{\bf Classification Accuracy} & \multirow{2}{*}{\bf Ranking} \\
\cline{5-8}
 &  & & & {Service A} & {Service B} & {Service C} & {Service D} & \\
 \cline{3-8}
\hline
\hline
benign & 16762 & 0.04 & 0.05 & 0.761 & 0.815 & 0.661 & 0.853 & DBAC \\
\hline
NSFW & 5276 & 0.14 & 0.05 & 0.965 & 0.804 & 0.533 & 0.767 & ABDC \\
\hline
malware & 1913 & 0.38 & 0.8 & 0.890 & 0.845 & 0.602 & 0.872 & ADBC \\
\hline
phishing & 1675 & 0.44 & 0.1 & 0.968 & 0.811 & 0.521 & 0.771 & ABDC \\
\hline
\multicolumn{4}{|c|}{Accuracy} & {\bf 0.826} & { 0.815} & 0.621 & {\bf 0.831} & DABC \\
\hline
\multicolumn{4}{|c|}{Balanced Accuracy} & {\bf 0.896} & { 0.819} & 0.579 & {\bf 0.816} & ABDC \\
\hline
\multicolumn{4}{|c|}{\bf \systemrare} & {\bf 0.929} & {\bf 0.823} & 0.559 & {\bf 0.812} & ABDC \\
\hline
\multicolumn{4}{|c|}{\bf \systemuser} & {\bf 0.895} & {\bf 0.838} & 0.593 & {\bf 0.856} & ADBC \\
\hline
\end{tabular}
\end{table*}

\begin{table*}[tb]
\caption{Training and evaluating a URLNet model using WBA}
\label{tab:url-eval}
\vspace{-2mm}
\centering
\scriptsize
\begin{tabular}{|l|r|r||r|r|r|r|r|}
\hline
\multirow{2}{*}{\bf Category} & \multicolumn{4}{|c|}{\bf Dataset Statistics} & \multicolumn{3}{|c|}{\bf Classification Accuracy} \\
\cline{2-8}
 & {\bf Train \#URLs} & {\bf Test \#URLs} & {\bf Rarity $w_i$} & {\bf User $w_i$} & {\bf Train with no {\system}} & {\bf Train with rarity $w_i$} & {\bf Train with user $w_i$} \\
\cline{2-8} 
\hline
\hline
benign & 10000 & 6762 & 0.04 & 0.05 & 0.981 & 0.300 & 0.458 \\
\hline
NSFW & 3150 & 2126 & 0.14 & 0.15 & {\bf 0} & {\bf 0.608} & 0.418 \\
\hline
malware & 1143 & 770 & 0.38 & 0.45 & {\bf 0.705} & {\bf 0.839} & {\bf 0.895} \\
\hline
phishing & 1000 & 675 & 0.44 & 0.35 &  0.782 & 0.788 & 0.754 \\
\hline
\multicolumn{5}{|c|}{\bf Test with Accuracy} & {0.745} & { 0.435} & {0.502} \\
\hline
\multicolumn{5}{|c|}{\bf Test with Balanced Accuracy} & { 0.617} & {0.634} & {0.631} \\
\hline
\multicolumn{5}{|c|}{\bf Test with \systemrare} & {\bf 0.653} & {\bf 0.761} & {NA} \\
\hline
\multicolumn{5}{|c|}{\bf Test with \systemuser} & {\bf 0.640} & {NA} & {\bf 0.752} \\
\hline
\end{tabular}
\vspace{-3mm}
\end{table*}

\vspace{-0.1cm}
\subsection{Use Case 3: URL Classification} 
\label{sec:url-exp}
\vspace{-0.1cm}


URL classification is a crucial task in the cyber-security industry for managing web traffic. Given a URL, the goal is to categorize the corresponding webpage into several well-defined classes such as benign (i.e., no harmful content), malware, phishing, NSFW (i.e., not safe for work), etc.
Given the vast number of URLs available, accurately learning and evaluating a URL classifier can greatly help in automatically categorizing webpages without requiring manual labeling \citep{vallina:2020:acm}. \\
\textbf{Dataset.}
Our dataset contains URLs from several different types of categories: benign (e.g., \emph{news}, \emph{sports}), NSFW (e.g., \emph{drugs}, \emph{gambling}), and malicious (\emph{phishing} and \emph{malware}). These URLs were sampled from a variety of third-party sources so as to minimize potential data bias. Many of the phishing and malware URLs were sampled from \emph{VirusTotal} (\cite{virustotal}), which aggregates various cyber-security vendors' detections into a single site, while benign URLs were sampled from \emph{Alexa's} Top Sites lists (\cite{alexa}), which rank websites according to their popularity. 
This dataset is representative of the types of URLs encountered in real-world URL filtering scenarios.

\textbf{URL Classification Services.}
We tested the performance of URL filtering products from 4 commercial companies, whose names had to be anonymized as A, B, C, D.
Since each URL filtering product has its own unique taxonomy of categories, we created mappings from each service's category space to a single shared category space so that we can easily compare the results.
Furthermore, since no company had access to our test dataset 
sampled from various third-party sources, no company was able to gain an unfair advantage in this URL classification task. \\
\textbf{Results.}
We submit the sampled URLs from each category to the above services, calculate their per-category accuracy, and compute a final accuracy score for competitive analysis. \\
\underline{\emph{Evaluating and Ranking the Classifiers:}}
Table \ref{tab:url-metrics} shows our results for Companies A-D. For overall $Accuracy$, D is the best as it has the highest classification accuracy on benign class (i.e., the majority class).
With $BalancedAccuracy$, we can see that A is the best, since it has higher accuracy for non-benign classes compared with other services, and this advantage is more evident with \systemrare, shown as a larger gap between their \systemrare values. B is better than D when using \systemrare, but if users consider malicious class to be the most important and apply a very high weight on it (e.g., 0.8), the resulted \systemuser metric would show that D is preferable over B. \\
\underline{\emph{Improving Model Training with WBA:}}
In this experiment, we adopt our dataset to train a URLNet model~(\cite{le2018urlnet}) by applying various WBA weights as suggested in Section~\ref{sec:weight}, to show its effectiveness in terms of improving corresponding WBA metric. As shown in Table~\ref{tab:url-eval}, 60\% of the dataset is used for training while the remaining 40\% is for evaluation.
Table~\ref{tab:url-eval} Classification Accuracy columns show the results under different settings. For vanilla training (without class weights), we get a high accuracy for the benign class but low accuracy for the others.
The main reason is benign class has the most number of URLs in training data. In order to boost the accuracy for non-benign classes, we apply rarity weights. With this, the accuracy for all non-benign and rarer classes are improved (noticeably, from 0 to 0.608 for the NSFW category). Note that malware content is often considered more harmful among all non-benign categories. For the third experiment, we apply user-defined weights which assign higher weight on the malware class, successfully boosting its accuracy from 0.839 to 0.895. 
It is clear that applying corresponding class weights in training can significantly improve the related WBA accuracy on the test dataset. For instance, when applying rarity weights, \systemrare improves from 0.653 to 0.761; while \systemuser is improved from 0.640 to 0.752 if the same user-defined weights are used in training. This aligns with our observation that increasing the weights in concerned categories can notably improve the accuracy in those categories.
As a comparison, neither the Accuracy nor the Balanced Accuracy measure does a good job in ranking the different training methods.

%% file: conclusion.tex
\vspace{-0.15cm}
\section{Conclusion} \label{sec:conclusion}
\vspace{-0.15cm}

In this paper, we presented a simple yet general-purpose class-sensitive evaluation framework for imbalanced data classification.
Our framework is designed to improve the grading of multi-class classifiers in domains where class importance is not evenly distributed. We provided a modular and extensible formulation
that can be easily customized to different importance criteria and metrics. Experiments with three real-world use cases show the value of a metric based on our framework, \systemcap\  (\system), over existing metrics -- in not only evaluating the classifiers' test results more sensitively to importance criteria, but also training them so. Under this framework, an interesting future direction would be to explore principled methods to assign the weights objectively for different applications.

%% file: appendix.tex
\section{Appendix} \label{sec-appendix}

In this appendix, we provide details for the experimental study, including data and code. For further information, please see our open source repository, which can be found on GitHub at:

\url{https://github.com/2023-07-03/weighted-balanced-accuracy}

\subsection{Details for Log Parsing Experiments}
\label{sec:appendix-log}

For the three log parsing techniques used in Section \ref{sec:log-exp} (Drain, Spell, and MoLFI), we used the implementations provided by the LogPAI team:

\url{https://github.com/logpai/logparser/}

The four datasets used in these experiments (macOS, BGL, Android, and HDFS) came from the benchmarking data also provided by LogPAI:

\url{https://github.com/logpai/loghub/}

\begin{figure}[t]
    \begin{center}
    \subfloat[macOS (skew = 8.454)]{
    \includegraphics[width=\linewidth]{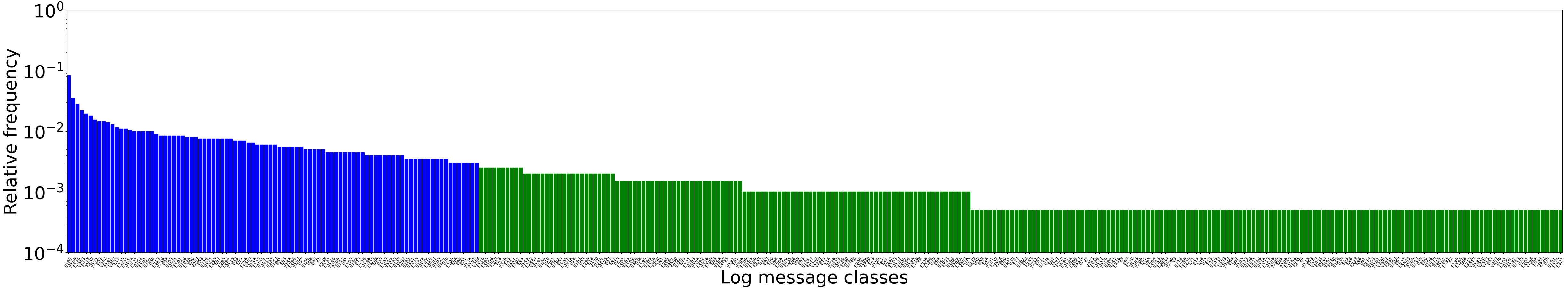}
    \label{fig:class-macos}
    }
    \newline
    \subfloat[BGL (skew = 8.900)]{
    \includegraphics[width=\linewidth]{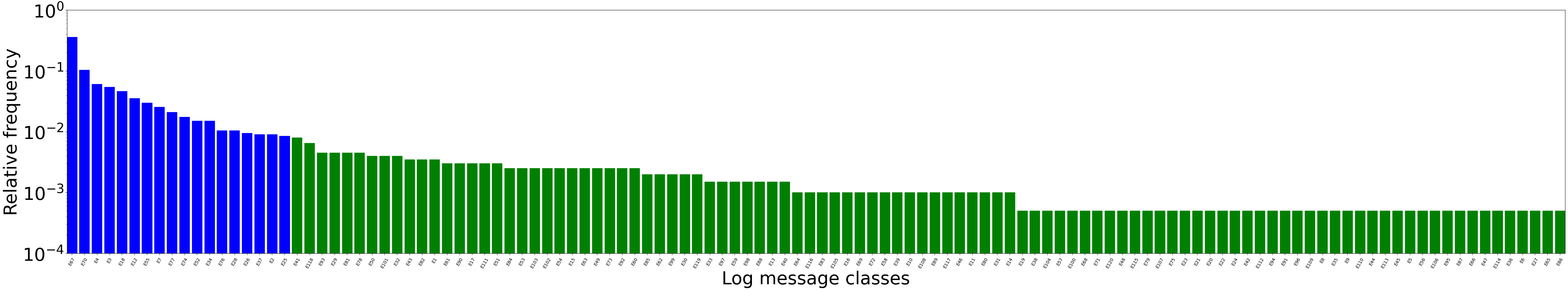}
    \label{fig:class-bgl}
    }
    \newline
    \subfloat[Android (skew = 4.822)]{
    \includegraphics[width=\linewidth]{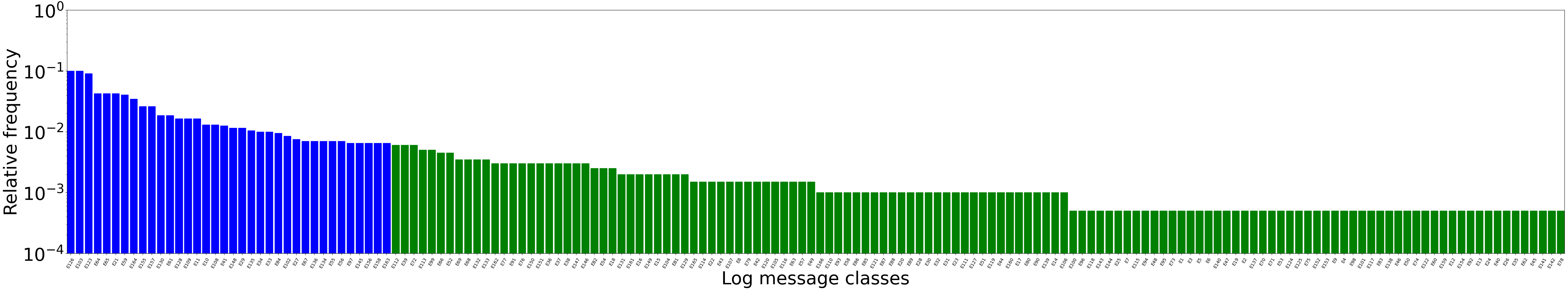}
    \label{fig:class-android}
    }
    \newline
    \subfloat[HDFS (skew = 0.202)]{
    \includegraphics[width=\linewidth]{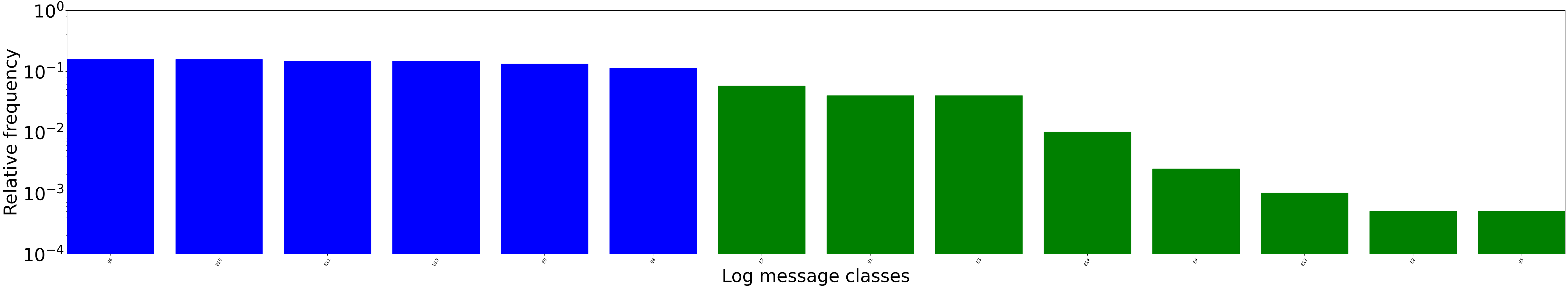}
    \label{fig:class-hdfs}
    }
    \caption{Histograms showing the relative frequencies of log parsing classes for the four experimental datasets: All graphs have their y-axes in log scale; green bars show the {\em infrequent classes}.}
    \label{fig:appendix-classdist-log}
    \end{center}
\end{figure}

In Figure \ref{fig:appendix-classdist-log}, we show the histograms for the four log datasets together with their skew values. As defined in the Microsoft Excel Documentation, ``{\em Skewness characterizes the degree of asymmetry of a distribution around its mean. Positive skewness indicates a distribution with an asymmetric tail extending toward more positive values, while negative skewness indicates a distribution with an asymmetric tail extending toward more negative values.}"
\footnote{\scriptsize{
\url{https://support.microsoft.com/en-us/office/skew-function-bdf49d86-b1ef-4804-a046-28eaea69c9fa}}}. In our context, skew provides a good indication for the degree of imbalance in class cardinality distributions -- the larger the skew, the larger the degree of class imbalance.

We also provide data files with class labels (true + predicted) and weights (based on rarity as importance criteria) used in generating the experimental data plotted in Figure \ref{fig:exp-log} as part of our WBA-Evaluator tool implementation included in the supplementary material (can be found under the {\tt WBA-Evaluator/examples/LogParsing/}
directory).

\begin{figure}[t]
    \begin{center}
    \subfloat[Drain mis-classification count for macOS dataset]{
    \includegraphics[width=\linewidth]{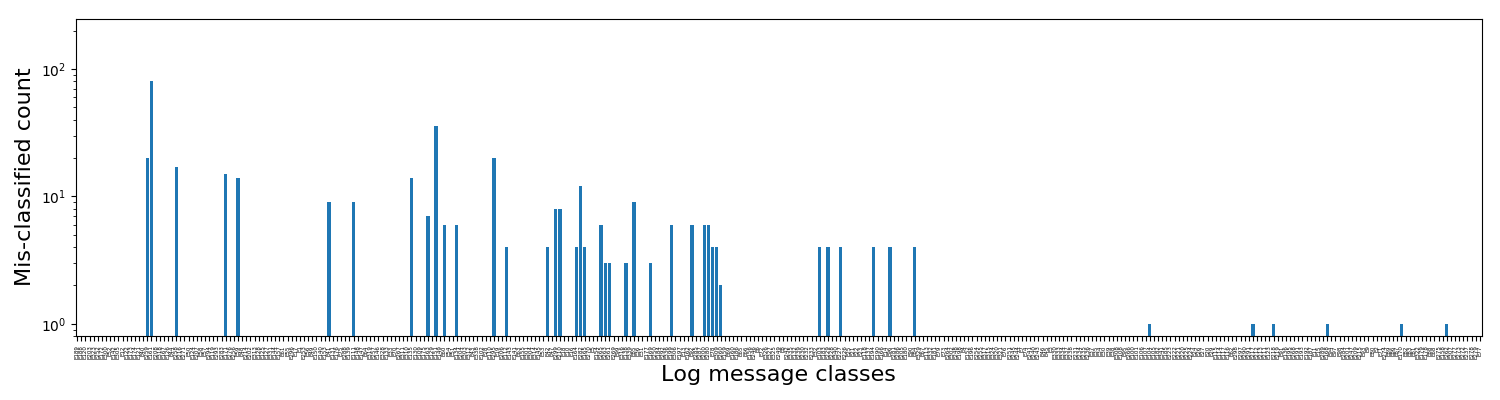}
    \label{fig:macos-drain}
    }
    \newline
    \subfloat[Spell mis-classification count for macOS dataset]{
    \includegraphics[width=\linewidth]{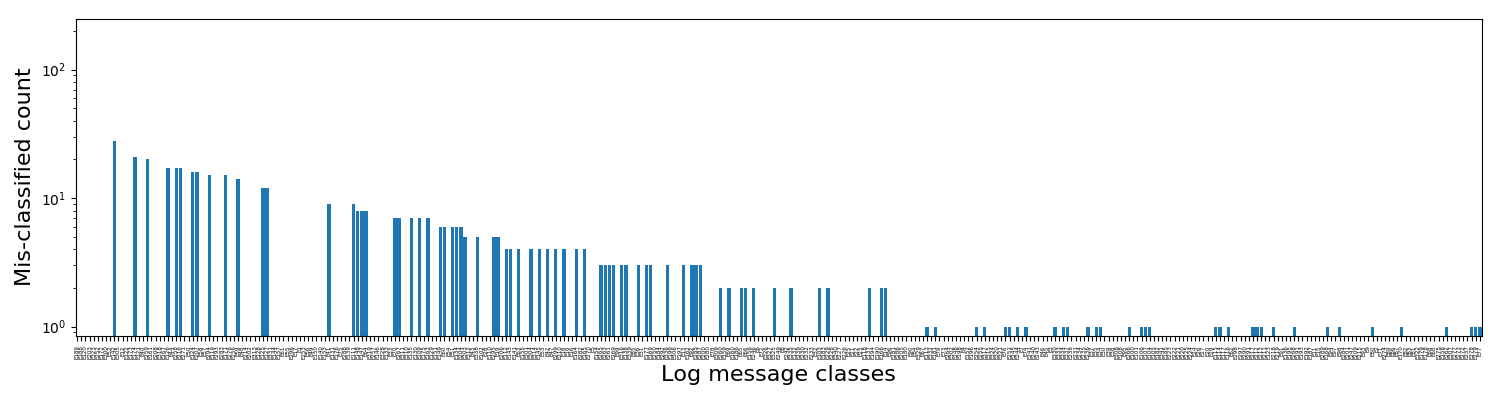}
    \label{fig:macos-spell}
    }
    \newline
    \subfloat[MoLFI mis-classification count for macOS dataset]{
    \includegraphics[width=\linewidth]{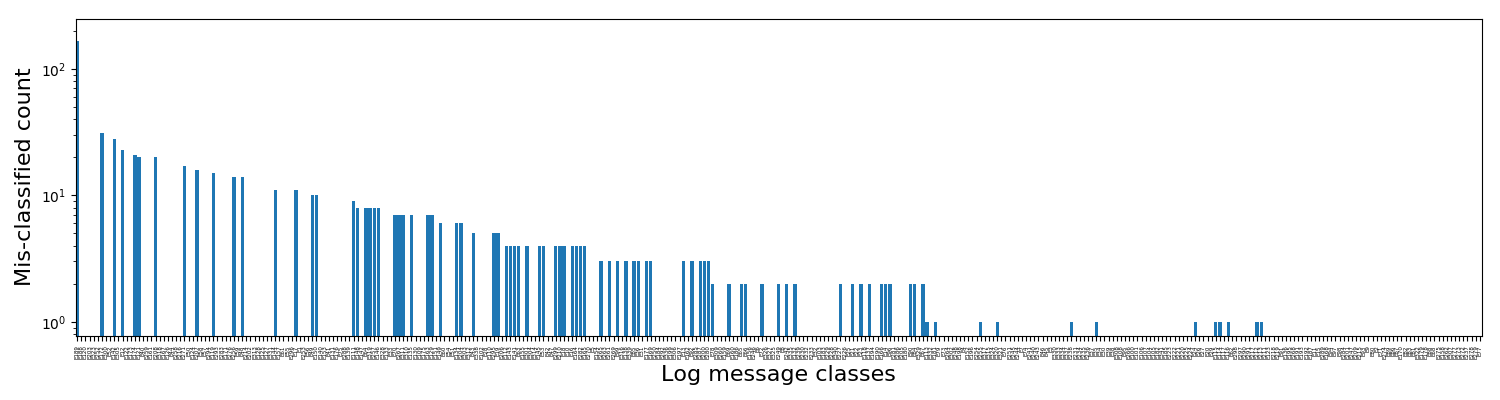}
    \label{fig:macos-molfi}
    }
    \caption{Misclassification count of three different algorithms on the macOS dataset. x-axis is ranked in descending order of class frequencies, as in Figure~\ref{fig:class-macos}. Drain performs best on rare classes (right side), which aligns with the {\systemrare} ranking shown in Figure~\ref{fig:res-macos}.}
    \label{fig:appendix-misclassification-macos}
    \end{center}
\end{figure}

\begin{figure}[t]
    \begin{center}
    \subfloat[Drain mis-classification count for BGL dataset]{
    \includegraphics[width=\linewidth]{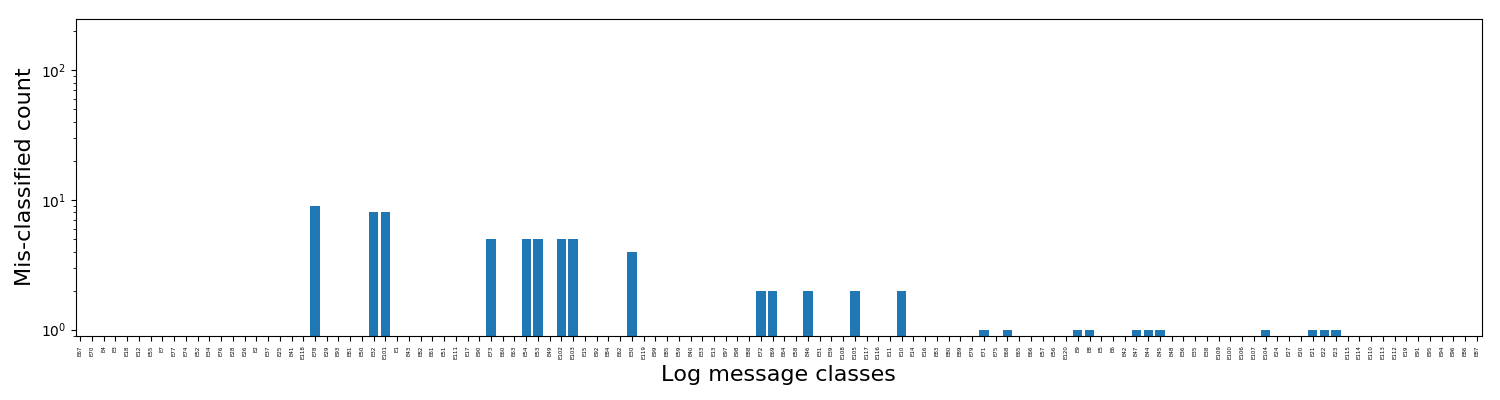}
    \label{fig:bgl-drain}
    }
    \newline
    \subfloat[Spell mis-classification count for BGL dataset]{
    \includegraphics[width=\linewidth]{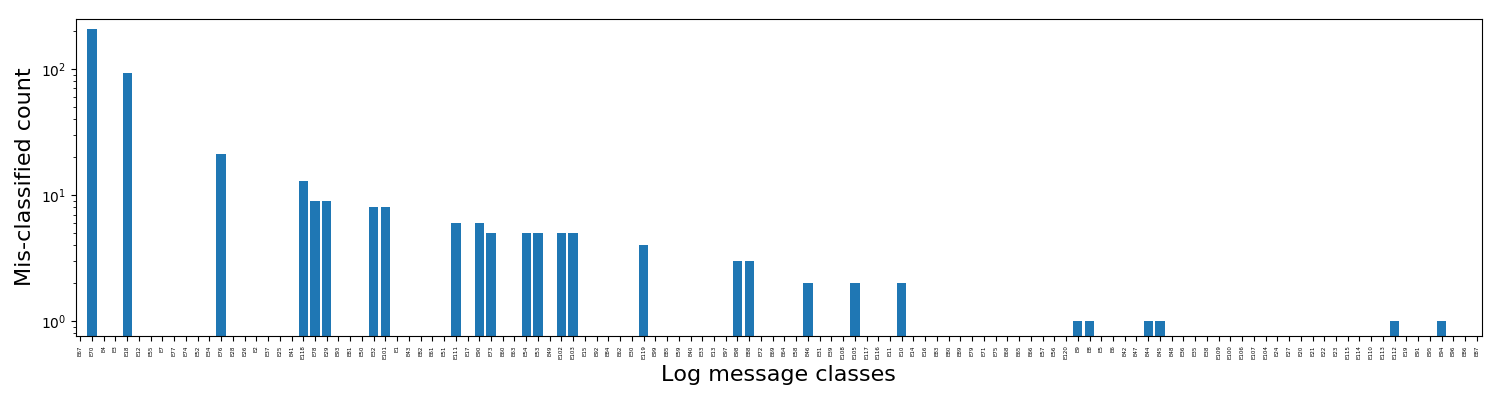}
    \label{fig:bgl-spell}
    }
    \newline
    \subfloat[MoLFI mis-classification count for BGL dataset]{
    \includegraphics[width=\linewidth]{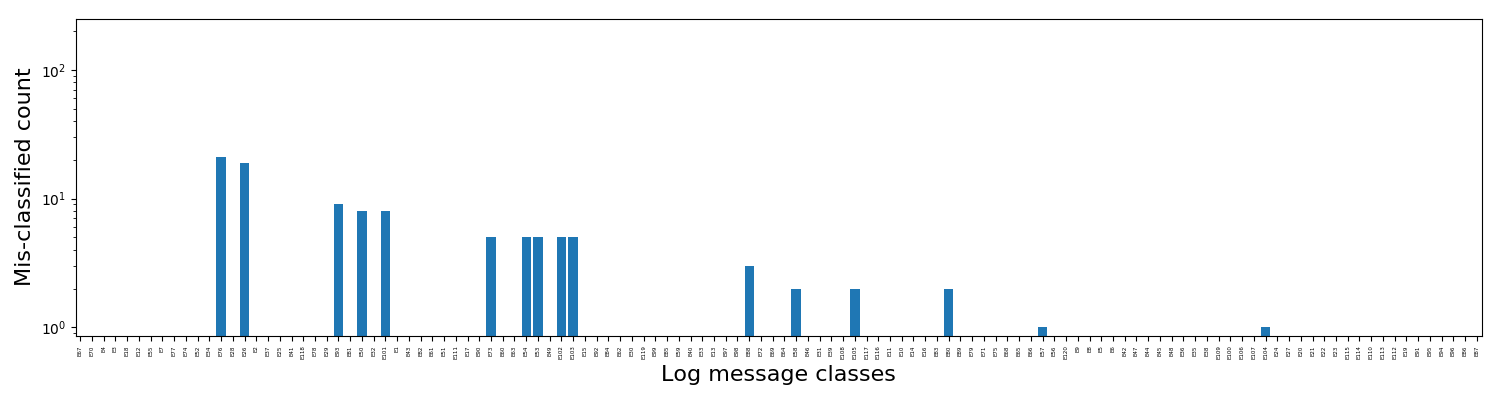}
    \label{fig:bgl-molfi}
    }
    \caption{Misclassification count of three different algorithms on the BGL dataset. x-axis is ranked in descending order of class frequencies, as in Figure~\ref{fig:class-bgl}.  MoLFI performs best on rare classes (right side), which aligns with the {\systemrare} ranking shown in Figure~\ref{fig:res-bgl}.}
    \label{fig:appendix-misclassification-bgl}
    \end{center}
\end{figure}

\begin{figure}[t]
    \begin{center}
    \subfloat[Drain mis-classification count for Android dataset]{
    \includegraphics[width=\linewidth]{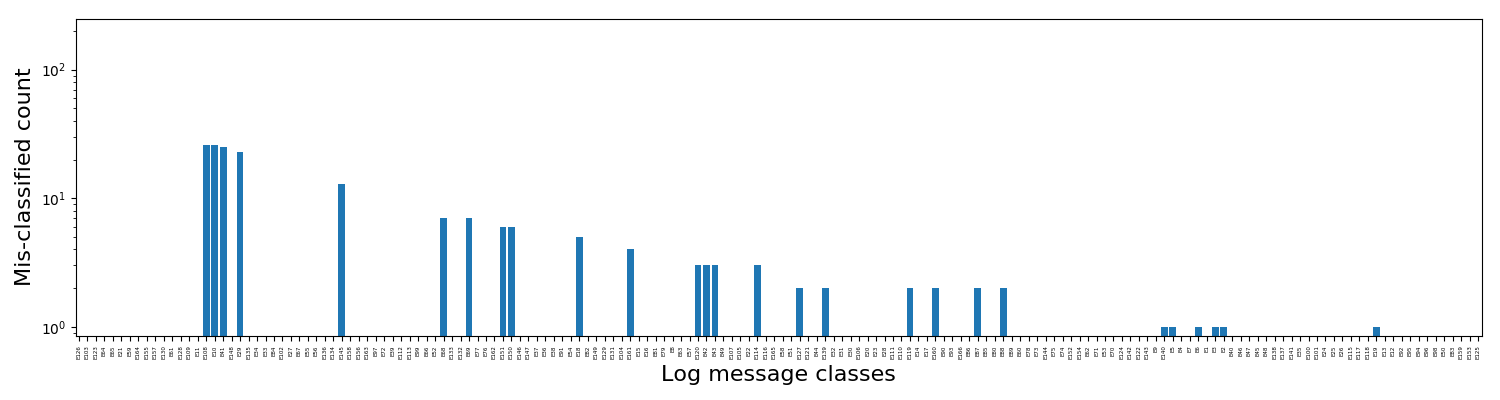}
    \label{fig:android-drain}
    }
    \newline
    \subfloat[Spell mis-classification count for Android dataset]{
    \includegraphics[width=\linewidth]{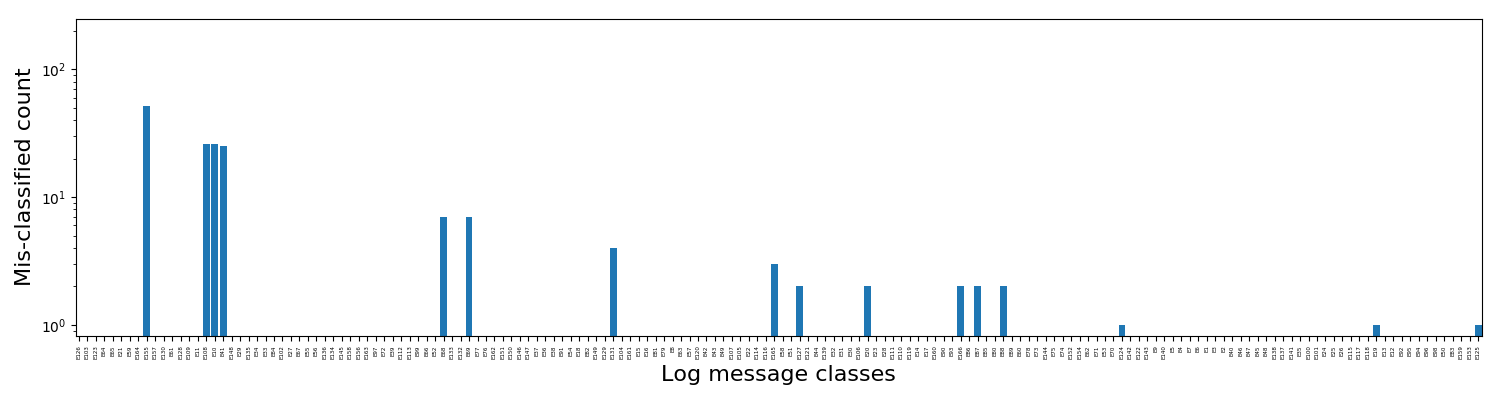}
    \label{fig:android-spell}
    }
    \newline
    \subfloat[MoLFI mis-classification count for Android dataset]{
    \includegraphics[width=\linewidth]{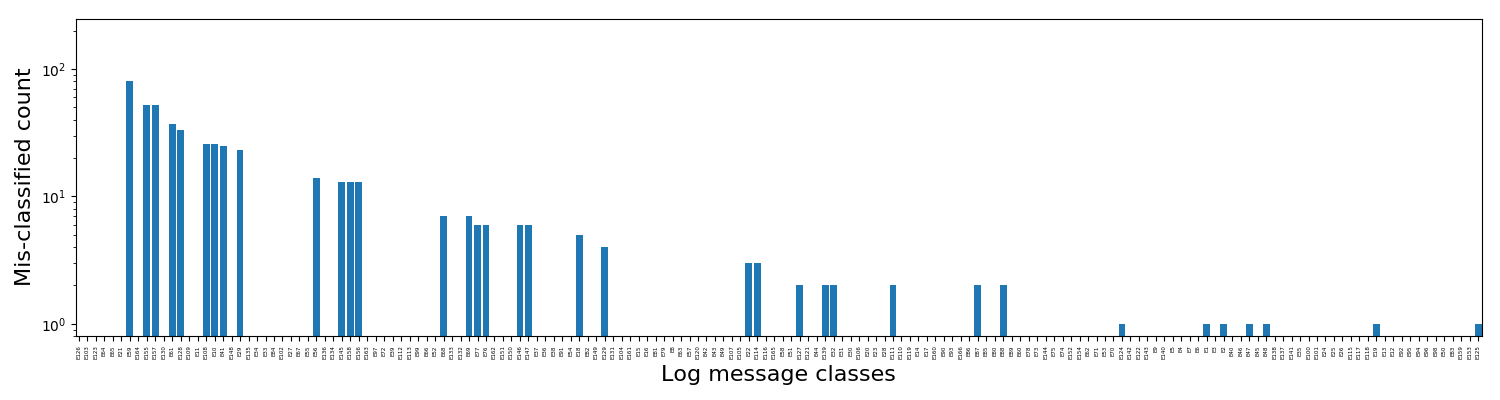}
    \label{fig:android-molfi}
    }
    \caption{Misclassification count of three different algorithms on the Android dataset. x-axis is ranked in descending order of class frequencies, as in Figure~\ref{fig:class-bgl}.  Spell performs best on rare classes (right side), which aligns with the {\systemrare} ranking shown in Figure~\ref{fig:res-android}.}
    \label{fig:appendix-misclassification-android}
    \end{center}
\end{figure}

A visualization of the per-class mis-classification count for each method on each dataset can be found from Figure \ref{fig:appendix-misclassification-macos} to Figure \ref{fig:appendix-misclassification-android}. The lower a bar is, the fewer samples being mis-classified in that category. Note that the categories on x axis are ordered in a descending order of class frequencies, i.e., in the same order of Figure~\ref{fig:appendix-classdist-log}. As a result, the lower the bars on the right side of each plot, the better the log parser is in classifying infrequent classes for that dataset, and thus the better the \systemrare score should be. As explained by the legends in each figure, the per-class accuracy in each plot aligns with {\systemrare} metrics in Figure~\ref{fig:exp-log}.

\subsection{Details for Sentiment Analysis Experiments}

For the sentiment analysis experiments of Section \ref{subsec:amazon}, we used a sample from the Amazon Customer Reviews dataset provided at:

\url{https://nijianmo.github.io/amazon/index.html}

\begin{figure}[t]
    \begin{center}
    \includegraphics[width=0.5\linewidth]{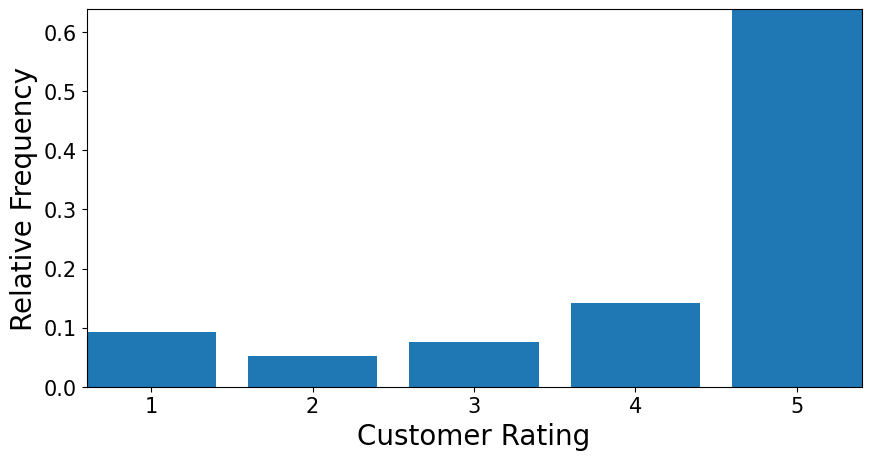}
    \caption{Histogram showing the relative frequencies of the five customer rating classes for the Amazon dataset (skew = 2.140).}
    \label{fig:appendix-classdist-amazon}
    \end{center}
\end{figure}

In Figure \ref{fig:appendix-classdist-amazon}, we show the histogram for the Amazon dataset. As described in Section \ref{subsec:amazon}, we implemented 4 RNN-based classifiers to experiment with this dataset. The code for these classifiers can be found in the supplementary material (under the {\tt AmazonReviewsClassifier/src/} directory) along with a copy of the data (under the {\tt AmazonReviewsClassifier/dataset/} directory).

We also provide the data files with class labels (true + predicted) and weights (user) used in generating the experimental data for LSTM results plotted in Figure \ref{fig:exp-amazon} and Table \ref{tab:stats-amazon} as an example. These can be found in our WBA-Evaluator tool implementation included in the supplementary material under the {\tt WBA-Evaluator/examples/Amazon/} directory.

\subsection{Details for URL Classification Experiments}
For the URL classification experiment in Section~\ref{sec:url-exp}, and specifically for the experiment result in Table~\ref{tab:url-eval}, our implementation is inherited from the URLNet work (\cite{le2018urlnet}). In particular, we changed the URLNet source code to support multi-class classification (previously binary), and to apply various weights for model training. The URLNet open source repository is available at:

\url{https://github.com/Antimalweb/URLNet/}

Our implementation and modification can be found at the  {\tt WBA-Evaluator/WeightedURLNet/} directory.

In the main context, we show how the proposed WBA weights are able to improve training accuracy for underrepresented classes. Due to page limit, only the evaluation results on 4 classes are listed in Figure~\ref{tab:url-eval}. Here we include the evaluation results for 2 classes and 3 classes respectively, to further express the effectiveness of WBA weights in improving model training.
As shown in Table~\ref{tab:url-eval-2classes}, by applying {\systemrare} weights and user-defined weights which all focus more on phishing category, the phishing accuracy is improved accordingly in both cases, with corresponding {\system} metric improved as well. In contrast, neither overall accuracy nor Balanced Accuracy is able to show the accuracy improvement in the important class.
Similarly, Table~\ref{tab:url-eval-3classes} shows the experiment results for 3 classes. It should be noted that a vanilla training leads to 0 accuracy on the newly added NSFW category, while another training with rarity weights brings some accuracy to this category, and applying a relatively high user weight for this category boosts its accuracy to over 80\%.

\begin{table}[t]
\caption{Training and evaluating a URLNet model using WBA - 2 classes}
\label{tab:url-eval-2classes}
\centering
\scriptsize
\begin{tabular}{|l|r|r||r|r|r|r|r|}
\hline
\multirow{3}{*}{\bf Category} & \multicolumn{4}{|c|}{Dataset Statistics} & \multicolumn{3}{|c|}{Classification Accuracy} \\
\cline{2-8}
 & {\bf Train} & {\bf Test} & {\bf Rarity} & {\bf User} & {\bf Train with} & {\bf Train with } & {\bf Train with } \\
 & {\bf \#URLs} & {\bf \#URLs} & {\bf  $w_i$} & {\bf  $w_i$} & {\bf no {\system} } &  {\bf rarity $w_i$} &  {\bf user $w_i$} \\
\cline{2-8} 
\hline
\hline
Benign & 10000 & 6762 & 0.091 & 0.2 & 0.994 & 0.902 & 0.950 \\
\hline
Phishing & 1000 & 675 & 0.909 & 0.8 &  0.859 & 0.942 & 0.861 \\
\hline
\multicolumn{5}{|c|}{\bf Test with Accuracy} & {0.958} & { 0.903} & {0.942} \\
\hline
\multicolumn{5}{|c|}{\bf Test with Balanced Accuracy} & { 0.927} & {0.922} & {0.906} \\
\hline
\multicolumn{5}{|c|}{\bf Test with \systemrare} & {\bf 0.854} & {\bf 0.904} & {NA} \\
\hline
\multicolumn{5}{|c|}{\bf Test with \systemuser} & {\bf 0.868} & {NA} & {\bf 0.879} \\
\hline
\end{tabular}
\end{table}

\begin{table}[t]
\caption{Training and evaluating a URLNet model using WBA - 3 classes}
\label{tab:url-eval-3classes}
\centering
\scriptsize
\begin{tabular}{|l|r|r||r|r|r|r|r|}
\hline
\multirow{3}{*}{\bf Category} & \multicolumn{4}{|c|}{Dataset Statistics} & \multicolumn{3}{|c|}{Classification Accuracy} \\
\cline{2-8}
 & {\bf Train} & {\bf Test} & {\bf Rarity} & {\bf User} & {\bf Train with} & {\bf Train with } & {\bf Train with } \\
 & {\bf \#URLs} & {\bf \#URLs} & {\bf  $w_i$} & {\bf  $w_i$} & {\bf no {\system} } &  {\bf rarity $w_i$} &  {\bf user $w_i$} \\
\cline{2-8} 
\hline
\hline
Benign & 10000 & 6762 & 0.07 & 0.1 & 0.992  & 0.795  & 0.177 \\
\hline
NSFW & 3150 & 2126 & 0.22 & 0.5 & {\bf 0}  & {\bf 0.105}  & {\bf 0.832} \\
\hline
Phishing & 1000 & 675 & 0.71 & 0.4 & 0.855  & 0.933  & 0.901 \\
\hline
\multicolumn{5}{|c||}{\bf Test with Accuracy} & {\bf 0.762} & {\bf 0.652} & {0.374} \\
\hline
\multicolumn{5}{|c||}{\bf Test with Balanced Accuracy} & {0.616} & {0.611} & {0.637} \\ \hline
\multicolumn{5}{|c||}{\bf Test with \systemrare} & {\bf 0.673} & {\bf 0.738} & {NA} \\
\hline
\multicolumn{5}{|c||}{\bf Test with \systemuser} & {\bf 0.441} & {NA} & {\bf 0.794} \\
\hline
\end{tabular}
\end{table}

\subsection{The WBA-Evaluator Tool}

In addition to details on our experimental study as described above, we also provide a copy of the WBA-Evaluator tool that implements our customizable, skew-sensitive evaluation framework described in Section \ref{sec:system}. WBA-Evaluator is written in Python and can be found in the supplementary material along with a README that describes how it can be used. In a nutshell, WBA-Evaluator takes as input three files (true class labels, predicted class labels, class weights) and a number of configuration parameters in the form of commandline arguments, and then it generates accuracy scores (BA or WBA) as specified by these arguments. The WBA-Evaluator implementation comes with two subdirectories: {\tt src/} contains the Python source code; {\tt example/} contains all the input files (labels and weights) and scripts in the {\tt scripts/} subfolder to run these. Please see the {\tt README} file for more details. Using this tool, results reported in the paper can be reproduced.


%% file: main_dmlr2023.bbl
\begin{thebibliography}{39}
\providecommand{\natexlab}[1]{#1}
\providecommand{\url}[1]{\texttt{#1}}
\expandafter\ifx\csname urlstyle\endcsname\relax
  \providecommand{\doi}[1]{doi: #1}\else
  \providecommand{\doi}{doi: \begingroup \urlstyle{rm}\Url}\fi

\bibitem[Sci()]{Scikit}
{Scikit-learn Classification Metrics}.
\newblock
  \url{https://scikit-learn.org/stable/modules/model_evaluation.html#classification-metrics}.

\bibitem[Amazon()]{alexa}
Amazon.
\newblock {Alexa Top Sites}.
\newblock \url{https://www.alexa.com/topsites}.

\bibitem[Batuwita \& Palade(2012)Batuwita and Palade]{batuwita:jbcb12}
Batuwita, R. and Palade, V.
\newblock {Adjusted Geometric-mean: A Novel Performance Measure for Imbalanced
  Bioinformatics Datasets Learning}.
\newblock \emph{{Journal of Bioinformatics and Computational Biology}},
  10\penalty0 (4), 2012.

\bibitem[Blaszczynski \& Stefanowski(2015)Blaszczynski and
  Stefanowski]{blaszczynski15}
Blaszczynski, J. and Stefanowski, J.
\newblock {Neighbourhood Sampling in Bagging for Imbalanced Data}.
\newblock \emph{{Neurocomputing}}, 150:\penalty0 529--542, 2015.

\bibitem[Branco et~al.(2016)Branco, Torgo, and Ribeiro]{branco:2016:survey}
Branco, P., Torgo, L., and Ribeiro, R.~P.
\newblock {A Survey of Predictive Modeling on Imbalanced Domains}.
\newblock \emph{ACM Computing Surveys (CSUR)}, 49\penalty0 (2):\penalty0 1--50,
  2016.

\bibitem[Brodersen et~al.(2010)Brodersen, Ong, Stephan, and
  Buhmann]{brodersen:2010:balanced}
Brodersen, K.~H., Ong, C.~S., Stephan, K.~E., and Buhmann, J.~M.
\newblock {The Balanced Accuracy and Its Posterior Distribution}.
\newblock In \emph{{20th International Conference on Pattern Recognition
  (ICPR)}}, pp.\  3121--3124, 2010.

\bibitem[Carrillo et~al.(2014)Carrillo, Brodersen, and
  Castellanos]{carrillo:2014:balanced}
Carrillo, H., Brodersen, K.~H., and Castellanos, J.~A.
\newblock {Probabilistic Performance Evaluation for Multi-class Classification
  using the Posterior Balanced Accuracy}.
\newblock In \emph{{First Iberian Robotics Conference}}, pp.\  347--361, 2014.

\bibitem[Castro \& de~P{\'{a}}dua~Braga(2013)Castro and
  de~P{\'{a}}dua~Braga]{castro13}
Castro, C.~L. and de~P{\'{a}}dua~Braga, A.
\newblock {Novel Cost-Sensitive Approach to Improve the Multilayer Perceptron
  Performance on Imbalanced Data}.
\newblock \emph{{IEEE Transactions on Neural Networks and Learning Systems}},
  24\penalty0 (6):\penalty0 888--899, 2013.

\bibitem[Cohen et~al.(2006)Cohen, ad~Hugo~Sax, Hugonnet, and
  Geissbuhler]{cohen:2006:cwa}
Cohen, G., ad~Hugo~Sax, M.~H., Hugonnet, S., and Geissbuhler, A.
\newblock {Learning from Imbalanced Data in Surveillance of Nosocomial
  Infection}.
\newblock \emph{{Artificial Intelligence in Medicine}}, 37\penalty0
  (1):\penalty0 7--18, 2006.

\bibitem[Du \& Li(2016)Du and Li]{du:2016:spell}
Du, M. and Li, F.
\newblock {Spell: Streaming Parsing of System Event Logs}.
\newblock In \emph{{IEEE International Conference on Data Mining (ICDM)}}, pp.\
   859--864, 2016.

\bibitem[Du \& Li(2018)Du and Li]{du:2018:spell}
Du, M. and Li, F.
\newblock {Spell: Online Streaming Parsing of Large Unstructured System Logs}.
\newblock \emph{{IEEE Transactions on Knowledge and Data Engineering (TKDE)}},
  31\penalty0 (11):\penalty0 2213--2227, 2018.

\bibitem[Estabrooks \& Japkowicz(2001)Estabrooks and Japkowicz]{estabrooks01}
Estabrooks, A. and Japkowicz, N.
\newblock {A Mixture-of-Experts Framework for Learning from Imbalanced Data
  Sets}.
\newblock In \emph{{International Conference on Advances in Intelligent Data
  Analysis (IDA)}}, pp.\  34--43, 2001.

\bibitem[Estabrooks et~al.(2004)Estabrooks, Jo, and Japkowicz]{estabrooks04}
Estabrooks, A., Jo, T., and Japkowicz, N.
\newblock {A Multiple Resampling Method for Learning from Imbalanced Data
  Sets}.
\newblock \emph{{Computational Intelligence}}, 20\penalty0 (1):\penalty0
  18--36, 2004.

\bibitem[He \& Garcia(2009)He and Garcia]{he:tkde09}
He, H. and Garcia, E.~A.
\newblock {Learning from Imbalanced Data}.
\newblock \emph{{IEEE Transactions on Knowledge and Data Engineering (TKDE)}},
  21\penalty0 (9):\penalty0 1263--1284, 2009.

\bibitem[He \& Ma(2013)He and Ma]{he13:book}
He, H. and Ma, Y. (eds.).
\newblock \emph{{Imbalanced Learning: Foundations, Algorithms, and
  Applications}}.
\newblock {Wiley-IEEE Press}, July 2013.

\bibitem[He et~al.(2017)He, Zhu, Zheng, and Lyu]{he:2017:drain}
He, P., Zhu, J., Zheng, Z., and Lyu, M.~R.
\newblock {Drain: An Online Log Parsing Approach with Fixed Depth Tree}.
\newblock In \emph{{IEEE International Conference on Web Services (ICWS)}},
  pp.\  33--40, 2017.

\bibitem[Helff et~al.(2016)Helff, Gruenwald, and d'Orazio]{helff16}
Helff, F., Gruenwald, L., and d'Orazio, L.
\newblock {Weighted Sum Model for Multi-Objective Query Optimization for
  Mobile-Cloud Database Environments}.
\newblock In \emph{{EDBT/ICDT International Workshop on Multi-Engine Data
  AnaLytics (MEDAL)}}, 2016.

\bibitem[Hilario et~al.(2018)Hilario, Lopez, Galar, Prati, Krawczyk, and
  Herrera]{hilario18:book}
Hilario, A.~F., Lopez, S.~G., Galar, M., Prati, R.~C., Krawczyk, B., and
  Herrera, F.
\newblock \emph{{Learning from Imbalanced Data Sets}}.
\newblock {Springer}, 2018.

\bibitem[Japkowicz(2013)]{japkowicz13}
Japkowicz, N.
\newblock {Assessment Metrics for Imbalanced Learning}.
\newblock In He, H. and Ma, Y. (eds.), \emph{{Imbalanced Learning: Foundations,
  Algorithms, and Applications}}, pp.\  187--206. {John Wiley \& Sons}, 2013.

\bibitem[Johnson \& Khoshgoftaar(2019)Johnson and Khoshgoftaar]{dl-survey19}
Johnson, J.~M. and Khoshgoftaar, T.~M.
\newblock {Survey on Deep Learning with Class Imbalance}.
\newblock \emph{{Journal of Big Data}}, 6\penalty0 (27), 2019.

\bibitem[Joshi et~al.(2001)Joshi, Kumar, and Agarwal]{joshi01}
Joshi, M.~V., Kumar, V., and Agarwal, R.~C.
\newblock {Evaluating Boosting Algorithms to Classify Rare Classes: Comparison
  and Improvements}.
\newblock In \emph{{IEEE International Conference on Data Mining (ICDM)}}, pp.\
   257--264, 2001.

\bibitem[Le et~al.(2018)Le, Pham, Sahoo, and Hoi]{le2018urlnet}
Le, H., Pham, Q., Sahoo, D., and Hoi, S.~C.
\newblock Urlnet: Learning a url representation with deep learning for
  malicious url detection.
\newblock \emph{arXiv preprint arXiv:1802.03162}, 2018.

\bibitem[Makki et~al.(2019)Makki, Assaghir, Taher, Haque, Hacid, and
  Zeineddine]{makki:access19}
Makki, S., Assaghir, Z., Taher, Y., Haque, R., Hacid, M., and Zeineddine, H.
\newblock {An Experimental Study With Imbalanced Classification Approaches for
  Credit Card Fraud Detection}.
\newblock \emph{{IEEE Access}}, 7:\penalty0 93010--93022, 2019.

\bibitem[Maloof(2003)]{maloof03}
Maloof, M.~A.
\newblock {Learning When Data Sets are Imbalanced and When Costs are Unequal
  and Unknown}.
\newblock In \emph{{ICML Workshop on Learning from Imbalanced Data Sets}},
  2003.

\bibitem[Messaoudi et~al.(2018)Messaoudi, Panichella, Bianculli, Briand, and
  Sasnauskas]{messaoudi:2018:molfi}
Messaoudi, S., Panichella, A., Bianculli, D., Briand, L., and Sasnauskas, R.
\newblock {A Search-based Approach for Accurate Identification of Log Message
  Formats}.
\newblock In \emph{{Proceedings of the 26th Conference on Program
  Comprehension}}, pp.\  167--177, 2018.

\bibitem[Mudambi \& Schuff(2010)Mudambi and Schuff]{mudambi:2010:amazon}
Mudambi, S.~M. and Schuff, D.
\newblock {Research Note: What Makes a Helpful Online Review? A Study of
  Customer Reviews on Amazon.com}.
\newblock \emph{{MIS Quarterly}}, 34\penalty0 (1):\penalty0 185--200, 2010.

\bibitem[Ni et~al.(2019)Ni, Li, and McAuley]{ni:2019:ijcnlp}
Ni, J., Li, J., and McAuley, J.
\newblock {Justifying Recommendations using Distantly-Labeled Reviews and
  Fine-Grained Aspects}.
\newblock In \emph{{Conference on Empirical Methods in Natural Language
  Processing and the 9th International Joint Conference on Natural Language
  Processing (EMNLP-IJCNLP)}}, pp.\  188--197, 2019.

\bibitem[Pavlou \& Dimoka(2006)Pavlou and Dimoka]{pavlou:2006:informs}
Pavlou, P.~A. and Dimoka, A.
\newblock {The Nature and Role of Feedback Text Comments in Online
  Marketplaces: Implications for Trust Building, Price Premiums, and Seller
  Differentiation}.
\newblock \emph{{Information Systems Research}}, 17\penalty0 (4):\penalty0
  392--414, 2006.

\bibitem[Pennington et~al.(2014)Pennington, Socher, and
  Manning]{pennington:2014:glove}
Pennington, J., Socher, R., and Manning, C.~D.
\newblock {GloVe: Global Vectors for Word Representation}.
\newblock In \emph{{Empirical Methods in Natural Language Processing (EMNLP)}},
  pp.\  1532--1543, 2014.

\bibitem[PyTorch()]{pytorch-class-weights}
PyTorch.
\newblock {PyTorch cross entropy loss API}.
\newblock
  \url{https://pytorch.org/docs/stable/generated/torch.nn.CrossEntropyLoss.html}.

\bibitem[Subramanian(2018)]{subramanian:2018:pytorch}
Subramanian, V.
\newblock \emph{{Deep Learning with PyTorch: A Practical Approach to Building
  Neural Network Models using PyTorch}}.
\newblock {Packt Publishing}, 2018.

\bibitem[Sun et~al.(2009)Sun, Wong, and Kamel]{sun:ijpr09}
Sun, Y., Wong, A. K.~C., and Kamel, M.~S.
\newblock {Classification of Imbalanced Data: A Review}.
\newblock \emph{{International Journal of Pattern Recognition and Artificial
  Intelligence}}, 23\penalty0 (4):\penalty0 687--719, 2009.

\bibitem[TensorFlow()]{tf-class-weights}
TensorFlow.
\newblock {TensorFlow class weights}.
\newblock
  \url{https://www.tensorflow.org/guide/keras/train_and_evaluate#class_weights}.

\bibitem[Tofallis(2014)]{tofallis14}
Tofallis, C.
\newblock {Add or Multiply? A Tutorial on Ranking and Choosing with Multiple
  Criteria}.
\newblock \emph{{INFORMS Transactions on Education}}, 14\penalty0 (3):\penalty0
  109--119, 2014.

\bibitem[Triantaphyllou(2000)]{mcdm00}
Triantaphyllou, E.
\newblock \emph{{Multi-criteria Decision Making Methods: A Comparative Study}}.
\newblock {Springer}, 2000.

\bibitem[Vallina et~al.(2020)Vallina, Pochat, Feal, Paraschiv, Gamba, Burke,
  Hohlfeld, Tapiador, and Vallina-Rodriguez]{vallina:2020:acm}
Vallina, P., Pochat, V., Feal, Ã., Paraschiv, M., Gamba, J., Burke, T.,
  Hohlfeld, O., Tapiador, J., and Vallina-Rodriguez, N.
\newblock Mis-shapes, mistakes, misfits: An analysis of domain classification
  services.
\newblock pp.\  598--618, 10 2020.
\newblock \doi{10.1145/3419394.3423660}.

\bibitem[VirusTotal()]{virustotal}
VirusTotal.
\newblock {VirusTotal Documentation}.
\newblock
  \url{https://support.virustotal.com/hc/en-us/categories/360000162878-Documentation}.

\bibitem[Weng \& Poon(2008)Weng and Poon]{wauc08}
Weng, C.~G. and Poon, J.
\newblock {A New Evaluation Measure for Imbalanced Datasets}.
\newblock In \emph{{Proceedings of the 7th Australasian Data Mining Conference
  (AusDM'08)}}, pp.\  27--32, 2008.

\bibitem[Zhu et~al.(2019)Zhu, He, Liu, He, Xie, Zheng, and Lyu]{zhu:2019:icse}
Zhu, J., He, S., Liu, J., He, P., Xie, Q., Zheng, Z., and Lyu, M.~R.
\newblock {Tools and Benchmarks for Automated Log Parsing}.
\newblock In \emph{{IEEE/ACM 41st International Conference on Software
  Engineering: Software Engineering in Practice (ICSE-SEIP)}}, pp.\  121--130,
  2019.

\end{thebibliography}
